%% file: main.tex
\documentclass[10pt,logo]{dobottechreport}

\usepackage[numbers,square,sort&compress]{natbib}
\input{math_commands.tex}

\usepackage{array}
\usepackage{multirow}
\usepackage{tcolorbox}
\usepackage{algorithm}
\usepackage{algpseudocode}
\usepackage{placeins}
\usepackage{flafter}
\usepackage{titletoc}
\usepackage{wrapfig}
\tcbuselibrary{breakable}

\renewcommand{\sfdefault}{lmss}

\titleformat{\section}
  {\large\bfseries\headingfont}
  {\color{dobotblue}\thesection}
  {0.5em}
  {#1}
\titleformat{name=\section,numberless}
  {\Large\bfseries\headingfont}
  {}
  {0pt}
  {#1}
\titleformat{\subsection}
  {\large\bfseries}
  {\color{dobotblue}\thesubsection}
  {0.5em}
  {#1}
\titleformat{\subsubsection}
  {\bfseries}
  {\color{dobotblue}\thesubsubsection}
  {0.5em}
  {#1}

\newcommand{\takeaway}[2]{%
  \begin{tcolorbox}[
    breakable,
    colback=dobotblue!4,
    colframe=dobotblue!45,
    boxrule=0.4pt,
    leftrule=2pt,
    arc=1pt,
    left=5pt,
    right=5pt,
    top=3pt,
    bottom=3pt,
    before skip=6pt,
    after skip=6pt]
  \textbf{\color{dobotblue}Takeaway #1.}\ #2
  \end{tcolorbox}%
}

\title{OPTS-TTPO: Enhancing Finite-Sample Policy-Gradient Learning with Tree Search}
\input{authors.tex}
\hypersetup{
  pdftitle={OPTS-TTPO: Enhancing Finite-Sample Policy-Gradient Learning with Tree Search},
  pdfsubject={On-policy tree search and policy-gradient learning},
  pdfkeywords={tree search, policy gradients, on-policy reinforcement learning, credit assignment}
}

\begin{document}
\maketitle

\input{sections/0_Abstract.tex}
\abscontent

\input{sections/1_Introduction_zheng.tex}
\input{sections/2_Related_Works.tex}
\input{sections/3_Methods.tex}
\input{sections/4_Experiments.tex}
\input{sections/5_Conclusion.tex}

\bibliographystyle{unsrtnat}
\bibliography{references}

\clearpage
\appendix
\startcontents[appendix]
\begingroup
\hypersetup{hidelinks}
\setcounter{tocdepth}{2}
\noindent{\LARGE\bfseries\color{dobotblue} Table of Contents}\par
\vspace{0.35em}
{\color{dobotblue}\hrule}
\vspace{0.45em}
\small
\titlecontents{section}
  [2.4em]
  {\addvspace{0.25em}\bfseries}
  {\contentslabel{1.8em}}
  {}
  {\hfill\contentspage}
\titlecontents{subsection}
  [5.2em]
  {}
  {\contentslabel{2.8em}}
  {}
  {\titlerule*[0.55em]{.}\contentspage}
\printcontents[appendix]{}{1}{}
\endgroup
\clearpage
\setcounter{figure}{0}
\setcounter{table}{0}
\setcounter{equation}{0}
\renewcommand{\thefigure}{\thesection.\arabic{figure}}
\renewcommand{\thetable}{\thesection.\arabic{table}}
\renewcommand{\theequation}{\thesection.\arabic{equation}}
\renewcommand{\theHfigure}{appendix.\thesection.\arabic{figure}}
\renewcommand{\theHtable}{appendix.\thesection.\arabic{table}}
\renewcommand{\theHequation}{appendix.\thesection.\arabic{equation}}
\input{sections/Appendix.tex}

\end{document}

%% file: math_commands.tex
\usepackage{amsmath,amsfonts,bm}

\def\eqref#1{equation~\ref{#1}}

\def\1{\bm{1}}

\DeclareMathAlphabet{\mathsfit}{\encodingdefault}{\sfdefault}{m}{sl}
\SetMathAlphabet{\mathsfit}{bold}{\encodingdefault}{\sfdefault}{bx}{n}



%% file: authors.tex
\author{%
  \textbf{Junyu Lu}\textsuperscript{1,*,\S}\quad
  \textbf{Shichao Weng}\textsuperscript{1,*}\quad
  \textbf{Zhiqiang Wang}\textsuperscript{1}\quad
  \textbf{Haojie Luo}\textsuperscript{1}\quad
  \textbf{Jingfan Zhang}\textsuperscript{2} \\
  \textbf{Yuhua Zhou}\textsuperscript{3}\quad
  \textbf{Cheng Du}\textsuperscript{2}\quad
  \textbf{Yuzhuo Zhang}\textsuperscript{4}\quad
  \textbf{Xi Li}\textsuperscript{1} \\
  \textbf{Jinwei Du}\textsuperscript{2}\quad
  \textbf{Tiancheng Feng}\textsuperscript{1,\dag}\quad
  \textbf{Chuan Xiao}\textsuperscript{5}\quad
  \textbf{Shuyuan Zheng}\textsuperscript{5,\ddag} \\
  {\normalfont\Affilfont \textsuperscript{1}Dobot Robotics\quad
   \textsuperscript{2}Independent Researcher} \\
  {\normalfont\Affilfont \textsuperscript{3}Zhejiang University\quad
   \textsuperscript{4}Fudan University\quad
   \textsuperscript{5}Osaka University} \\
  {\normalfont\Affilfont \textit{First-author contact:}
   \href{mailto:jun_yu_lu@163.com}{jun\_yu\_lu@163.com}}
}
\equalcontribution{Equal contribution.\\
  \textsuperscript{\S}\ Junyu Lu is a student at Beijing Institute of Technology, Zhuhai.
  This work was done at Dobot Robotics.}
\projectleader{Project leader.}
\correspondingauthor{Shuyuan Zheng
  (\href{mailto:zheng@ist.osaka-u.ac.jp}{zheng@ist.osaka-u.ac.jp})}
\hypersetup{pdfauthor={Junyu Lu, Shichao Weng, Zhiqiang Wang, Haojie Luo, Jingfan Zhang, Yuhua Zhou, Cheng Du, Yuzhou Zhang, Xi Li, Jinwei Du, Tiancheng Feng, Chuan Xiao, Shuyuan Zheng}}

%% file: sections/0_Abstract.tex
\begin{abstract}
The policy-gradient theorem expresses the exact gradient as an expectation under the current policy, but practical methods estimate it from finitely many on-policy trajectories. Rare high-return trajectories may be missing from the policy-gradient estimate. We study whether tree search can improve their coverage under a finite budget while controlling the induced gradient bias. We introduce On-Policy Parallel Tree Search (OPTS) and Tree Trajectory Policy Optimization (TTPO) through \emph{on-policy tree trajectories}, which sample new suffixes from the current policy at visited states. This requires no action-distribution correction, but branching changes state visitation. Our Branch Aggregation Lemma underpins TTPO: branch-weighted tree statistics recover chain expectations when branch choices and weights are fixed before outgoing transitions are sampled. OPTS selects expansion states using estimated performance differences. Under deterministic dynamics, exact values, and max-backup advantages, the induced search policy's expected return improves monotonically with the budget. Adaptive expansion violates the lemma's condition; we bound its gradient bias and show that max backup adds prefix credit to actions leading to better discovered suffixes. Against a finite chain reference, TTPG's measured bias stays near its no-branching level while NaivePG's bias grows from $0.1251$ to $0.4884$. Exact-value search is monotone and learned-critic summaries improve in aggregate. At matched budgets, reward- and value-guided OPTS improve correct-answer coverage and majority-vote accuracy over i.i.d. baselines. At matched branch counts, the coverage--bias diagnostic shows that OPTS + TTPG trades a modest bias increase for greater coverage relative to Fixed-branch + TTPG. Under matched interaction or rollout budgets, OPTS-TTPO improves MuJoCo tail returns over PPO by up to $28.6\%$, posts a 34--22--1 win--loss--tie record against PPO on Atari-57 under the last-100-log mean-return metric, and improves both micro-averaged \texttt{avg@32} and \texttt{pass@32} over PPO across all four Qwen3 models.
\end{abstract}

%% file: sections/1_Introduction_zheng.tex
\section{Introduction}
\label{sec:introduction}

The policy-gradient theorem expresses the exact gradient as an expectation over trajectories sampled from the current policy~\citep{sutton1999policy}. Practical algorithms such as PPO estimate this expectation using a finite batch of on-policy trajectories~\citep{ref_ppo}. Under independent sampling, each trajectory starts from the initial-state distribution and provides one sampled suffix from each visited state. Collecting more independent trajectories can improve the gradient estimate, but rare high-return trajectories may still go unobserved under a finite sampling budget. This motivates studying how to allocate the available budget to improve high-return trajectory coverage for policy-gradient learning.

This work targets a central finite-sample failure of policy-gradient learning: a finite sample of trajectories can miss rare, high-return trajectories. We use tree search to improve the chance of covering such trajectories while controlling the bias introduced by adaptive search. Tree search offers a way to use the available budget more selectively~\citep{browne2012survey}. Whereas every independent trajectory must start from the initial-state distribution, search can select an informative state on an already sampled prefix and draw fresh suffixes from that state. A critic-derived estimate can identify visited states with the greatest estimated room for improvement and concentrate new samples there. This mechanism changes where samples are collected, not how actions are generated: every new action is still drawn from the current policy. Therefore, the resulting suffixes remain on-policy in their conditional action distribution and require no action-distribution importance correction.

\textbf{Core challenges.}
Turning this intuition into a policy-gradient algorithm is not straightforward. First, \emph{how should the learner aggregate a tree?} Branching duplicates prefixes and gives different states different numbers of descendants, so treating every node as an ordinary sample over-represents heavily expanded regions and generally changes the policy-gradient target, even when the advantage estimate at each transition is unbiased. Second, \emph{where should a finite search budget be spent?} Uniform branching wastes the very advantage of search; the algorithm needs a policy-relative criterion that identifies the states at which resampling is most valuable. Third, useful allocation is necessarily adaptive: the search chooses where to expand after seeing sampled outcomes. Such posterior-dependent selection conflicts with the conditions required for unbiased tree aggregation. Thus, adaptive search creates a genuine tradeoff between improved trajectory coverage and additional estimation bias.

\textbf{Our solution.}
We propose OPTS--TTPO, our framework that combines two methods introduced in this work: On-Policy Parallel Tree Search (OPTS) and Tree Trajectory Policy Optimization (TTPO). To make tree data usable for policy-gradient learning, we first prove the Branch Aggregation Lemma: when branching decisions and normalized local weights depend only on prefix information, a branch-weighted tree statistic has the same expectation as its ordinary on-policy chain counterpart. We then derive TTPO from this lemma, weighting each transition by its derived branch weight in the clipped PPO objective so that heavily expanded regions are not over-counted. To decide where to search, we introduce OPTS, which uses a critic-derived, policy-relative performance difference to estimate the benefit of replacing a sampled suffix with one generated by the current policy. OPTS directs the finite budget toward the highest-scoring visited states and samples new on-policy suffixes from them in parallel; under deterministic dynamics, exact current-policy values, and max-backup advantages, we prove that its induced search policy improves monotonically as the budget grows. Finally, we characterize what changes when OPTS and TTPO are combined: posterior-adaptive expansion violates the lemma's prefix-measurability condition. We decompose the resulting bias into posterior trajectory reweighting and the additional prefix credit introduced by max backup, which propagates the value of better suffixes back to their prefixes, and derive bounds for both effects. In stochastic environments, we instead use mean backup to reduce the amplification of favorable noise.

In summary, our contributions are fourfold: \textbf{(1) Principled learning from trees.} We prove the Branch Aggregation Lemma and introduce TTPO, whose branch-weighted updates recover the ordinary on-policy expectation under the lemma's conditions (Section~\ref{sec:method-ttpo}). \textbf{(2) Performance-directed search.} We introduce OPTS, a performance-difference-guided budget-allocation method with a monotonic search-improvement guarantee under ideal conditions (Section~\ref{sec:method-opts}). \textbf{(3) An explicit account of adaptivity.} We decompose and bound the bias of learning from posterior-adaptive OPTS trees with TTPO into trajectory reweighting and max-backup prefix credit (Section~\ref{sec:method-loop}). \textbf{(4) Theory-backed cross-domain validation.} Controlled mechanism experiments validate the theoretical results underlying OPTS--TTPO. Under matched budgets, OPTS--TTPO improves MuJoCo tail return by $18.7\%$ on average, posts a 34--22--1 win--loss--tie record against PPO on Atari-57 under the last-100-log mean-return metric, and improves LLM micro-averaged \texttt{avg@32} over PPO by $0.72$--$2.32$ percentage points and \texttt{pass@32} by $0.44$--$2.33$ percentage points across four models (Section~\ref{sec:experiments}).

%% file: sections/2_Related_Works.tex
\section{Related Work}
\label{sec:related-work}

\noindent\textbf{Adaptive allocation and high-return discovery.} Finite-budget policy-gradient learning and high-return discovery have been studied through adaptive data collection. Robust On-Policy Sampling addresses finite-sample mismatch in policy evaluation, while PROPS extends adaptive sampling to policy-gradient control~\citep{zhong2022robust,corrado2023props}; Adaptive Experience Selection changes replay selection to reduce gradient-estimator variance~\citep{mohamad2020adaptive}. Go-Explore and Self-Imitation Learning reuse promising states or past high-return trajectories~\citep{ecoffet2021goexplore,oh2018selfimitation}. For inference-time high-return discovery, repeated sampling provides a simple independent-sampling baseline~\citep{brown2024languagemonkeys}; recent RLVR methods operate at different allocation levels: VIP allocates rollouts across prompts, PROS reuses historical prefixes, and TRACE allocates additional suffixes at intermediate prefixes~\citep{nguyen2026adaptive,huang2026pros,zou2026trace}. These methods allocate computation at different units and for different targets; we focus on high-return trajectory coverage under a fixed rollout budget and its policy-gradient interface.

\noindent\textbf{Tree search as budget allocation.} Tree search uses posterior information to allocate search computation. MCTS formalizes selection, expansion, simulation, and backup~\citep{kocsis2006bandit,browne2012survey}; AlphaGo combines learned policy and value networks with MCTS for action selection~\citep{ref_alphago}. AlphaGo Zero, AlphaZero, MuZero, Expert Iteration, and related methods instead distill search-improved action distributions or trajectories into a policy~\citep{ref_alphagozero,ref_alphazero,ref_muzero,ref_exit,hubert2021learning,danihelka2022policy,grill2020mcts,ref_alphazerollm}. The tree is therefore commonly used to choose a root action, construct a search or execution policy, or provide a policy-projection target.

\noindent\textbf{Tree-structured policy-gradient learning.} Tree-structured policy-gradient methods use branches for value estimation or credit assignment. Tree MDPs, VinePPO, and Branching Policy Optimization use whole-tree, suffix, or sibling returns~\citep{scavuzzo2022treemdp,kazemnejad2024vineppo,he2026branching}; TreeRL, SPO-tree, TreeRPO, TreePO, and Tree-GRPO variants derive process or subtree signals from descendant outcomes~\citep{ref_treerl,guo2026segment,yang2025treerpo,ref_treepo,ref_treegrpo,ding2026treegrpo}. Other methods choose branches with uncertainty, lookahead, or information gain, or optimize search-internal policies and tree backups~\citep{ref_arpo,xing2026latr,zong2026at2po,wu2026spark,zhang2026igrpo,graf2016adaptive,zhang2019ace,farquhar2018treeqn,morimura2024policy,dalal2025policytreeexpansion,soemers2019learning,ref_seear1}. Their branching signals and learning targets differ: for example, TreeRL samples on-policy suffixes but uses group-relative process credit, while other methods optimize search-internal policies, persistent-tree policies, or subtree-derived objectives.

OPTS-TTPO targets finite-budget high-return trajectory coverage. OPTS allocates current-policy suffix rollouts to visited states using a policy-relative performance-difference estimate; TTPO learns from the resulting tree with branch-weighted aggregation, preserving the chain target under prefix-measurable expansion. We further analyze adaptive-selection bias and max-backup prefix credit.

%% file: sections/3_Methods.tex
\section{OPTS-TTPO}

\begin{figure}[t]
    \centering
    \includegraphics[width=\linewidth]{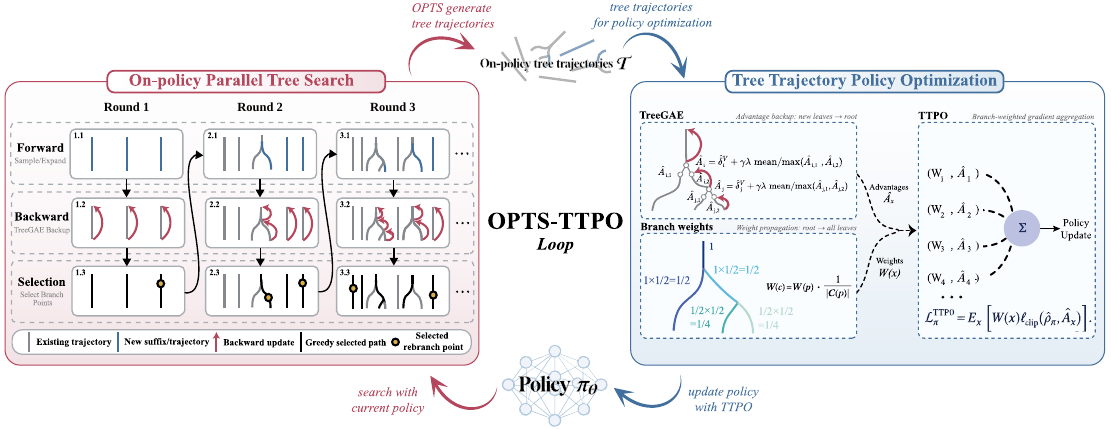}
    \caption{OPTS-TTPO. OPTS selects states maximizing the length-penalized performance-difference estimate and samples on-policy suffixes from them in parallel; TTPO applies TreeGAE and branch weights to PPO updates.}
    \label{fig:opts-ttpo-overview}
\end{figure}

OPTS-TTPO uses on-policy tree trajectories as the common interface between search and learning (Figure~\ref{fig:opts-ttpo-overview}). Branch aggregation is the foundation for learning from prefix-measurable trees, while performance-difference-guided OPTS allocates the finite rollout budget. We characterize the prefix credit and gradient bias that arise when posterior-adaptive OPTS is combined with TTPO. Complete definitions and proofs are given in Appendices~\ref{app:ttpo-theory}--\ref{app:opts-ttpo-theory}.

\noindent\textbf{On-policy tree trajectories.} An on-policy tree trajectory $\mathcal T=(s_o,\mathcal X(\mathcal T))$ consists of a root state $s_o\sim\rho_0$ and a finite rooted set of transition occurrences. Starting from $s_o$, it recursively samples one or more transitions according to $\pi_\theta$ and extends their successor states to the horizon; suffixes from a common state occurrence share the prefix ending there. Each transition $c$ from parent occurrence $p$ is $(s_p,a_c,r_c,s_c)$, with $a_c\sim\pi_\theta(\cdot\mid s_p)$ and $(r_c,s_c)\sim P(\cdot\mid s_p,a_c)$, and has a unique path from $s_o$. For transition index $x$, we write $(s_x,a_x,r_x,s_{x^+})$. Here, \emph{on-policy} refers only to the conditional action distribution $\pi_\theta(\cdot\mid s_p)$: every action occurrence in the tree is sampled from the current policy, so attaching suffixes requires no action-distribution importance correction. Branching nevertheless changes transition multiplicities relative to a single chain. The Branch Aggregation Lemma below derives weights that recover the chain-trajectory target under prefix-measurable branching.

\subsection{TTPO: How Should the Learner Aggregate a Tree?}
\label{sec:method-ttpo}

\noindent\textbf{Branch Aggregation Lemma.}\label{sec:branch-aggregation-lemma} Policy-gradient and GAE targets are chain-trajectory sums, whereas a tree can attach several suffixes to one shared prefix. As a simple example, if three suffixes branch at depth $t$, naive summation counts the suffix three times relative to the prefix; averaging the three suffixes restores the chain proportions. The lemma generalizes this correction to arbitrary trees.

For a horizon-$n$ chain $\tau=(x_0,\ldots,x_{n-1})$, define $H_\eta(\tau):=\sum_{t=0}^{n-1}\eta^t h(x_t)$ with scalar $\eta$ and integrable transition contribution $h$. The target is $\mathbb E_{\tau\sim\pi_\theta}[H_\eta(\tau)]$.

For each parent occurrence $p$, let its child set $\mathcal C(p)$ and local weights $\alpha_{p,c}\ge0$, with $\sum_{c\in\mathcal C(p)}\alpha_{p,c}=1$, be determined from prefix information. Define $W(o)=1$ and $W(c)=W(p)\alpha_{p,c}$. This prefix-measurability condition excludes dependence on the outgoing transitions or their descendants.

Let $d(x)$ be the transition depth, starting at zero. Direct aggregation sums the weighted contribution of every transition occurrence in the tree; recursive aggregation computes the discounted statistic of the subtree rooted at $s_p$ by averaging child contributions with their local weights, with zero at terminal:
\begin{equation}
\widehat H_{\eta,\mathrm{dir}}(\mathcal T)
:=
\sum_{x\in\mathcal T}W(x)\eta^{d(x)}h(x),
\qquad
\widehat R_\eta(s_p)
:=
\sum_{c\in\mathcal C(p)}\alpha_{p,c}
\bigl[h(c)+\eta\widehat R_\eta(s_c)\bigr].
\label{eq:main-branch-aggregation-forms}
\end{equation}
Under this prefix-measurability condition, the Branch Aggregation Lemma (Appendix~\ref{app:branch-aggregation}) gives $\widehat H_{\eta,\mathrm{dir}}(\mathcal T)=\widehat R_\eta(s_o)$ and
\begin{equation}
\mathbb E_{\mathcal T}
\left[\widehat H_{\eta,\mathrm{dir}}(\mathcal T)\right]
=
\mathbb E_{\tau\sim\pi_\theta}
\left[H_\eta(\tau)\right].
\label{eq:main-branch-aggregation}
\end{equation}

\noindent\textbf{Tree Trajectory Policy Gradient (TTPG).} Setting $\eta=\gamma$ and $h(x)=A^{\pi_\theta}(s_x,a_x)\nabla_\theta\log\pi_\theta(a_x\mid s_x)$ in the direct aggregation form gives
\begin{equation}
\nabla_\theta J(\theta)
=
\mathbb E_{\mathcal T}
\left[
\sum_{x\in\mathcal T}
W(x)\gamma^{d(x)}A^{\pi_\theta}(s_x,a_x)
\nabla_\theta\log\pi_\theta(a_x\mid s_x)
\right].
\label{eq:tree-policy-gradient}
\end{equation}
The weights $W(x)$ correct the state-visitation multiplicity introduced by branching.

\noindent\textbf{Tree-based Generalized Advantage Estimation (TreeGAE).} Let $\delta_x^V:=r_x+\gamma V(s_{x^+})-V(s_x)$. Setting $\eta=\gamma\lambda$ and $h(x)=\delta_x^V$ in the recursive aggregation form gives
\begin{equation}
\widehat A_x^{\mathrm{TreeGAE}(V)}
:=
\delta_x^V
+
\gamma\lambda
\sum_{x^+\in\mathcal C(x)}
\alpha_{x,x^+}\widehat A_{x^+}^{\mathrm{TreeGAE}(V)}.
\label{eq:main-tree-gae}
\end{equation}
TreeGAE and chain GAE have the same conditional suffix expectation; with $V=V^{\pi_\theta}$, Equation~\ref{eq:tree-policy-gradient} preserves the policy-gradient identity.

\noindent\textbf{Tree Trajectory Policy Optimization (TTPO).}\label{sec:ttpo-objective} Let $\rho_x(\theta):=\pi_\theta(a_x\mid s_x)/\pi_{\theta_{\mathrm{old}}}(a_x\mid s_x)$. TTPO applies the branch weights to the standard clipped PPO surrogate,
\begin{equation}
\mathcal L_\pi^{\mathrm{TTPO}}(\theta)
:=
\mathbb E_x
\left[
W(x)
\min\!\left(
\rho_x(\theta)\widehat A_x,
\operatorname{clip}\!\left(\rho_x(\theta),1-\epsilon,1+\epsilon\right)\widehat A_x
\right)
\right].
\label{eq:ttpo-actor}
\end{equation}

\subsection{OPTS: Where Should a Finite Search Budget Be Spent?}
\label{sec:method-opts}

\noindent\textbf{Performance-Difference Estimation.} Let $\tau=(s_t,a_t,r_t,\ldots,s_n)\sim(\mu,P)$ be a recorded suffix under an arbitrary policy $\mu$. Replacing its action at $s_k$ by a fresh action from the current policy $\pi$ and then following $\pi$ has expected local improvement $V^\pi(s_k)-Q^\pi(s_k,a_k)=-A^\pi(s_k,a_k)$. Their discounted sum defines the \emph{performance-difference estimate}, whose expectation follows from the performance difference lemma~\citep{kakade2002approximately,schulman2015trust}:
\begin{equation}
\Delta(s_t;\tau)
:=
-\sum_{k=t}^{n-1}\gamma^{k-t}A^\pi(s_k,a_k),
\qquad
\mathbb E_{\tau\sim(\mu,P)\mid s_t}[\Delta(s_t;\tau)]
=
V^\pi(s_t)-V^\mu(s_t).
\label{eq:main-perf-diff}
\end{equation}
In deterministic environments, $\Delta(s_t;\tau)=V^\pi(s_t)-G_t$ for recorded suffix return $G_t$. In practice, replacing $A^\pi$ by TreeGAE gives $\widehat\Delta(s_t;\tau):=-\sum_{k=t}^{n-1}\gamma^{k-t}\widehat A_{x_k}$, and we introduce the length penalty $\widehat\Delta^{(\xi)}(s_t;\tau):=\widehat\Delta(s_t;\tau)/(n-t)^\xi$, $\xi\in[0,1]$. When $\xi=0$, this recovers the original performance-difference estimate; when $\xi=1$, it averages the discounted local improvements along the suffix. Appendix~\ref{app:budget-alignment} gives further details.

\noindent\textbf{On-Policy Parallel Tree Search.} OPTS initializes on-policy chains and their TreeGAE advantages. Each round follows $c^\star(p)\in\arg\max_{c\in\mathcal C(p)}\widehat A_c$ to form an observed greedy path $\tau^\star$, then selects its highest-scoring candidate position:
\begin{equation}
s^\star\in\arg\max_{s_t\in\tau^\star}
\widehat\Delta^{(\xi)}(s_t;\tau^\star).
\label{eq:main-rebranch-state}
\end{equation}
A single batched rollout under the current policy generates suffixes from selected states and fresh trajectories from new initial states; TreeGAE is then updated from the new leaves back to the roots. A tree stops searching when it reaches $S_{\max}$ expansions or when no candidate exceeds a nonnegative baseline. Algorithm~\ref{alg:opts-search} gives the complete procedure.

In deterministic environments, OPTS uses max-backup TreeGAE:
\[
\widehat A_{\max}(x)
:=\delta_x^V+\gamma\lambda
\max_{c\in\mathcal C(x)}\widehat A_{\max}(c),
\qquad \max\varnothing:=0.
\]
The greedy paths of OPTS induce a search policy $\pi_j^S$, with $\pi_0^S=\pi$ (Appendix~\ref{app:opts-construction}). With exact $V=V^\pi$ and max-backup TreeGAE in a deterministic environment, OPTS satisfies
\[
J(\pi^S_{j_2})\ge J(\pi^S_{j_1})\ge J(\pi),
\qquad 0\le j_1\le j_2\le S_{\max}.
\]
Appendix~\ref{app:opts-improvement} also establishes monotonicity of the suffix $\lambda$-return objective under nested expansion.

\subsection{OPTS-TTPO: What Are the Benefits and Costs of the Adaptive Allocation?}
\label{sec:method-loop}

At iteration $u$, OPTS samples each suffix from the current policy but selects expansion states from observed outcomes, violating the lemma's prefix-measurability condition; TTPO then learns from the branch-weighted policy gradient of tree trajectories, $\pi_u\xrightarrow{\mathrm{OPTS}}\mathcal T_u\xrightarrow{\mathrm{TTPO}}\pi_{u+1}$.

In deterministic environments, OPTS-TTPO uses max-backup TreeGAE, which yields the prefix-credit term below; in stochastic environments, it uses mean-backup TreeGAE to average child advantages while retaining branch coverage. Both variants use $\alpha_{p,c}=1/|\mathcal C(p)|$ to correct branch multiplicity.

\noindent\textbf{Max-backup search information as prefix credit.} On a fixed tree, let $\widehat g_{\max}$ and $\widehat g_{\mathrm{mean}}$ be the TTPG directions under the two backups, and define the best-minus-average gap $b(x):=\max_{c\in\mathcal C(x)}\widehat A_{\max}(c)-\sum_{c\in\mathcal C(x)}\alpha_{x,c}\widehat A_{\max}(c)\ge0$, with $b(x)=0$ at leaves. Writing $x\preceq u$ when $x$ lies on the root-to-$u$ path, Appendix~\ref{app:posterior-prefix-credit} proves
\begin{align}
\widehat g_{\max}
&=\widehat g_{\mathrm{mean}}
+\left.\nabla_\theta\mathcal L_{\mathrm{search}}(\theta)\right|_{\theta=\theta_{\mathrm{old}}},
\nonumber\\
\mathcal L_{\mathrm{search}}(\theta)
&:=\gamma\lambda\sum_{u\in\mathcal T}W(u)\gamma^{d(u)}b(u)
\sum_{x\preceq u}\lambda^{d(u)-d(x)}\log\pi_\theta(a_x\mid s_x).
\label{eq:main-prefix-credit}
\end{align}
$\mathcal L_{\mathrm{search}}$ credits actions leading to each branch point by its gap $b(u)$, with the credit decaying by $\lambda$ at each step, so a discovered suffix can influence upstream learning before being reliably reproduced.

\noindent\textbf{Controlling the search-induced bias.} Let $p$ be the fraction of root trees receiving an expansion and $\rho_S\le p(1-2^{-S_{\max}})$ bound the expected path weight moved off the initial trajectory. Under bounded rewards, values, and policy scores, Appendix~\ref{app:search-gradient-pg-bias} gives
\begin{equation}
\|\overline g_{\max}-\nabla J(\theta_{\mathrm{old}})\|
\le
\varepsilon_0+
\min\left\{
2pC,\;
2\rho_SC+\varepsilon_{\mathrm{backup}}
\right\},
\label{eq:main-search-bias-bound}
\end{equation}
Here $C$ bounds tree and chain gradient norms, $\varepsilon_0$ is the initial chain-estimator bias (zero with exact values), and $\varepsilon_{\mathrm{backup}}$ is bounded by the weighted gap mass in Equation~\ref{eq:main-prefix-credit}. The coarse bound scales with $p$; the refined bound separates the posterior path reweighting shared by both backup rules, $2\rho_SC$, from the additional max-backup term $\varepsilon_{\mathrm{backup}}$. Algorithm~\ref{alg:opts-ttpo} gives the complete training step.

%% file: sections/4_Experiments.tex
\section{Experiments}
\label{sec:experiments}

We evaluate branch aggregation, OPTS search and test-time scaling, coverage--bias trade-offs, and cross-domain policy-gradient learning (Sections~\ref{sec:tree-policy-gradient-evaluation}--\ref{sec:cross-domain}). Rollout budgets are matched in LLM experiments and environment steps in control experiments; learned policies are evaluated without search. Full protocols and hyperparameters are in Appendix~\ref{app:experiment-setup}.

\begin{figure}[!t]
\centering
\includegraphics[width=0.85\linewidth]{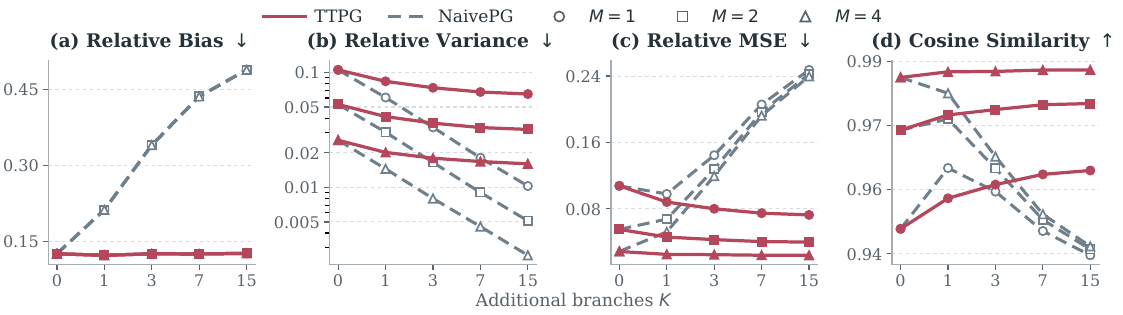}
\caption{\textbf{NaivePG versus TTPG} under global token-level normalization.}
\label{fig:rq1-gradient-aggregation}
\end{figure}

\begin{figure*}[!t]
\centering
\includegraphics[width=0.88\textwidth]{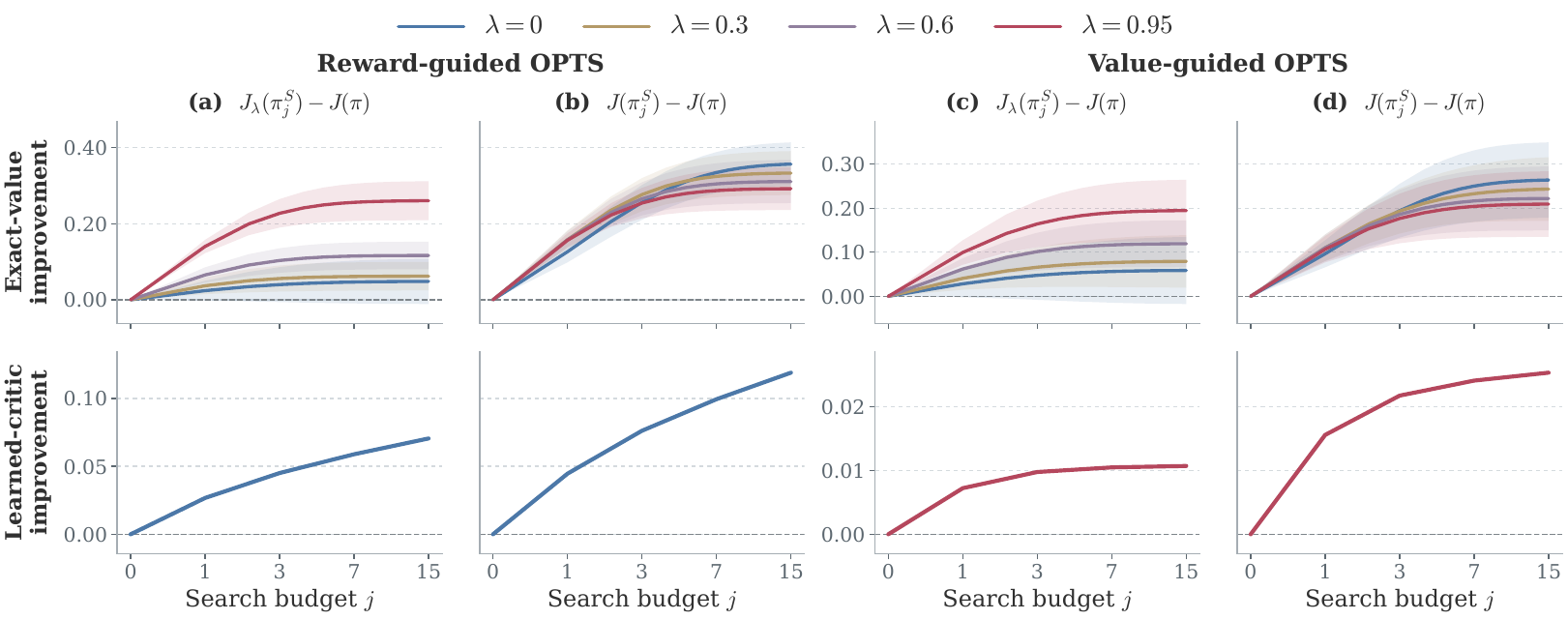}
\caption{Search-budget improvements. Top: exact-value; bottom: learned critic.}
\label{fig:opts-search-budget-evidence}
\end{figure*}

\subsection{Branch Aggregation for Policy-Gradient Estimation}
\label{sec:tree-policy-gradient-evaluation}

\noindent\textbf{Setup.} Using a frozen step-400 \texttt{Qwen3-1.7B} PPO checkpoint, we use unbiased zero-baseline advantage estimates from binary Monte Carlo returns, isolating branch aggregation from critic error. We compare NaivePG with TTPG on identical trees against a finite reference pooled from 32 independently sampled chain groups, varying additional suffixes $K\in\{0,1,3,7,15\}$ and tree groups $M\in\{1,2,4\}$ (Appendix~\ref{app:rq1-gradient-protocol}).

\noindent\textbf{Results.} At $K=0$ the two estimators coincide with relative bias $0.1251$ (Figure~\ref{fig:rq1-gradient-aggregation}). As suffixes are added, NaivePG bias grows monotonically to $0.4884$ at $K=15$, while TTPG stays between $0.1221$ and $0.1260$. Although reusing descendant tokens lowers NaivePG's variance, its squared bias dominates: at $M=4$ and $K=15$, NaivePG's relative MSE rises to $0.2398$ whereas TTPG's falls from $0.0285$ to $0.0239$, with cosine similarities of $0.9467$ and $0.9880$, respectively. Both estimators use the same unbiased advantage estimates; their divergence reflects NaivePG's over-counting of branched suffixes relative to shared prefixes. Branch weighting corrects this imbalance.

\takeaway{1}{Against the finite chain reference, NaivePG's measured bias grows with branching, whereas TTPG stays at its $K=0$ level and attains lower MSE and higher gradient alignment. Unbiased advantage estimates alone do not guarantee unbiased gradient estimates.}

\subsection{OPTS Search Quality and Test-Time Scaling}
\label{sec:opts-search-scaling}

\noindent\textbf{Exact-value improvement and learned-critic trends.} We test Theorem~2 on 32 depth-4 deterministic binary-trees with exact $V^\pi$, max backup, and zero baseline (Appendix~\ref{app:setup-rq2}). Reward- and value-guided OPTS use full and truncated advantages, respectively. Exact enumeration across $\lambda\in\{0,0.3,0.6,0.95\}$ shows that all 1,920 adjacent-budget comparisons per mode improve both expected $\lambda$-return and true return (Figure~\ref{fig:opts-search-budget-evidence}, top).

The learned-critic counterpart fixes a step-400 \texttt{Qwen3-1.7B} OPTS-TTPO policy and critic, uses $\lambda=0.999$, and varies $S_{\max}\in\{0,1,3,7,15\}$. At $S_{\max}=15$, the micro-averaged improvements in $(J_\lambda,J)$ are $(0.0706,0.1190)$ for reward-guided OPTS and $(0.0107,0.0254)$ for value-guided OPTS (Figure~\ref{fig:opts-search-budget-evidence}, bottom). All four pooled summaries increase with the search budget, although the gains are not uniformly monotone on every benchmark (Appendix~\ref{app:search-budget-improvement}). The gain also depends on the rebranching position (Appendix~\ref{app:rebranch-position-ablation}).

\begin{wrapfigure}{r}{0.56\textwidth}
\vspace{-0.8\baselineskip}
\centering
\includegraphics[width=0.54\textwidth]{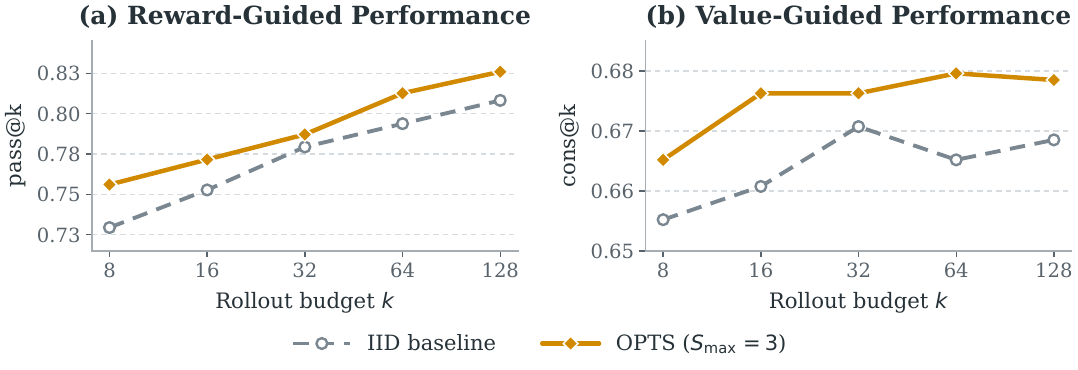}
\caption{\textbf{Matched-budget coverage and accuracy.} OPTS uses fixed $S_{\max}=3$; the IID curve is the matched baseline.}
\label{fig:rq2-compute-scaling}
\end{wrapfigure}

\noindent\textbf{Coverage and test-time scaling at matched rollout budgets.}\label{sec:llm-test-search} We evaluate OPTS with fixed $S_{\max}=3$ against independent sampling at matched rollout budgets $k\in\{8,16,32,64,128\}$, pooling results over 902 prompts; per-benchmark curves are in Appendix~\ref{app:rq2-compute-scaling-by-dataset}.

With verifier rewards guiding rebranching, we report whether a searched tree contains a correct answer; the matched baseline is \texttt{pass@k}. Reward-guided OPTS rises from $0.7561$ at $k=8$ to $0.8259$ at $k=128$, compared with $0.7295$ and $0.8082$ for independent sampling (Figure~\ref{fig:rq2-compute-scaling}(a)).

Without verifier rewards, OPTS uses the learned value function and truncated GAE to guide rebranching and returns the majority answer among the value-greedy tree responses; the baseline is \texttt{cons@}$k$, the majority-vote accuracy of $k$ i.i.d. samples~\citep{wang2022self}. Value-guided OPTS rises from $0.6652$ at $k=8$ to $0.6785$ at $k=128$ (peaking at $0.6796$ at $k=64$), around the corresponding IID values $0.6552$ and $0.6685$ (Figure~\ref{fig:rq2-compute-scaling}(b)).

\takeaway{2}{Larger search budgets improve $J_\lambda$ and $J$ under exact values, with the same aggregate trend under learned critics; at matched budgets, reward-guided and value-guided OPTS improve coverage and majority-vote accuracy, respectively, over independent sampling.}

\subsection{Coverage--Bias Trade-offs and Prefix Credit}
\label{sec:mechanistic-exp}

\begin{wrapfigure}{r}{0.56\textwidth}
\vspace{-0.8\baselineskip}
\centering
\includegraphics[width=0.54\textwidth]{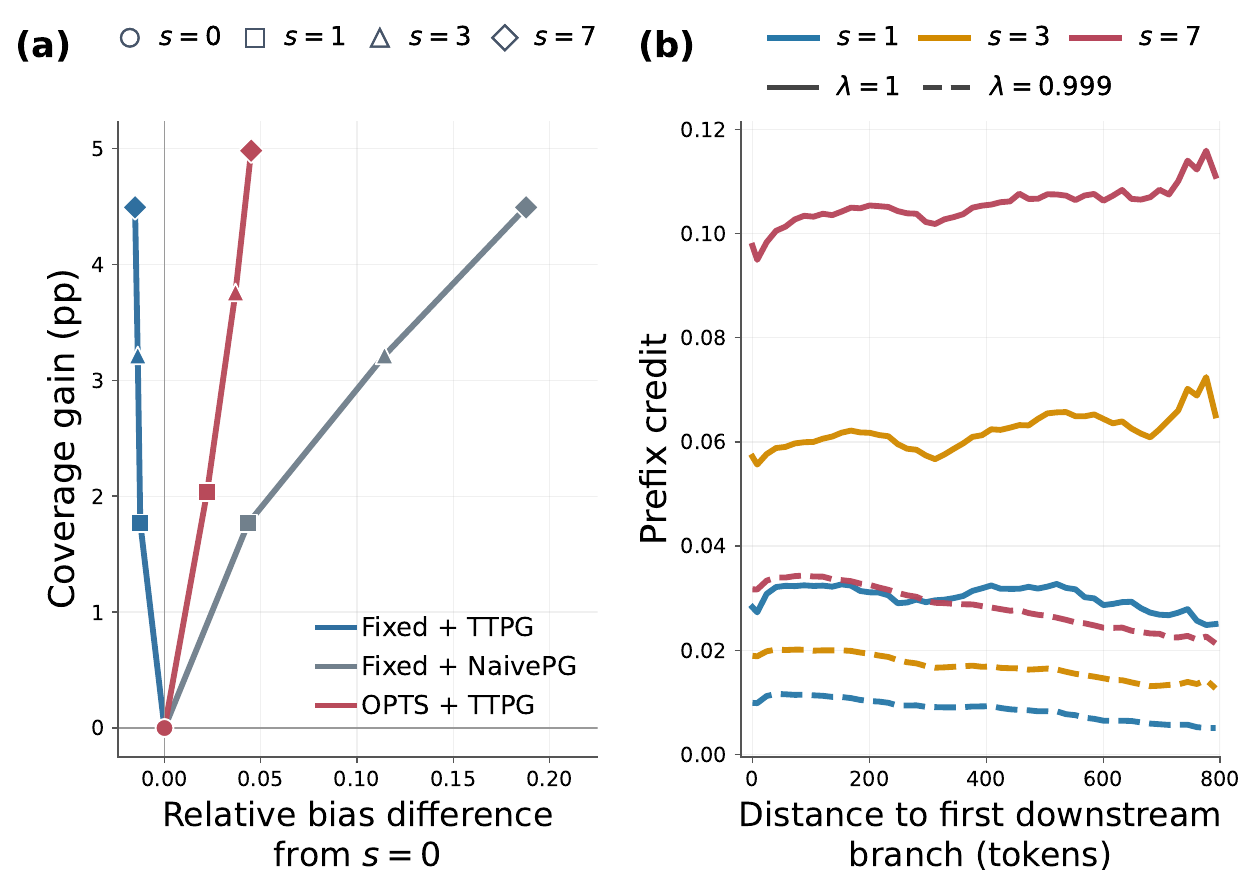}
\caption{\textbf{Mechanistic diagnostics.} (a) Coverage gain versus relative bias difference from $s=0$. (b) Token-weighted extra prefix credit (max$-$mean) versus distance to the first downstream branch.}
\label{fig:e1e2-mechanism-summary}
\end{wrapfigure}

Using a frozen step-400 \texttt{Qwen3-1.7B} OPTS-TTPO actor--critic checkpoint distinct from the PPO checkpoint above, we evaluate 16,384 training prompts. Fixed branching occurs at response token 128 and skips shorter responses; the other method uses OPTS. We compare $s\in\{0,1,3,7\}$ with $p=1$. Both panels use $M=1$ aggregation and direct returns with $V=0$; the right panel additionally compares max and mean backup.

At matched branch counts, Fixed-branch + TTPG gains coverage with a near-zero, slightly negative bias shift, whereas OPTS + TTPG attains greater coverage at the cost of a modest positive bias shift. Fixed-branch + NaivePG instead moves steadily toward larger positive bias. Panel~(b) shows that max backup assigns more prefix credit after more search rounds ($s=7>s=3>s=1$). With $\lambda=0.999$, this credit decays with distance from the first downstream branch; the $\lambda=1$ curves retain the credit over the plotted distances. Together, the panels illustrate the coverage--bias relationship and the additional prefix credit introduced by max backup.

\takeaway{3}{At matched branch counts, OPTS + TTPG trades a modest bias increase for greater coverage than Fixed-branch + TTPG, while max backup adds additional prefix credit.}

\subsection{Cross-Domain Policy-Gradient Learning}
\label{sec:cross-domain}

We compare OPTS-TTPO with PPO in three domains that differ in observation modality, action space, and budget unit. MuJoCo and Atari follow the CleanRL style~\citep{ref_cleanrl}; deterministic MuJoCo and LLM use max-backup TreeGAE, while sticky-action Atari uses mean backup.

\begin{figure}[!t]
\centering
\includegraphics[width=\linewidth]{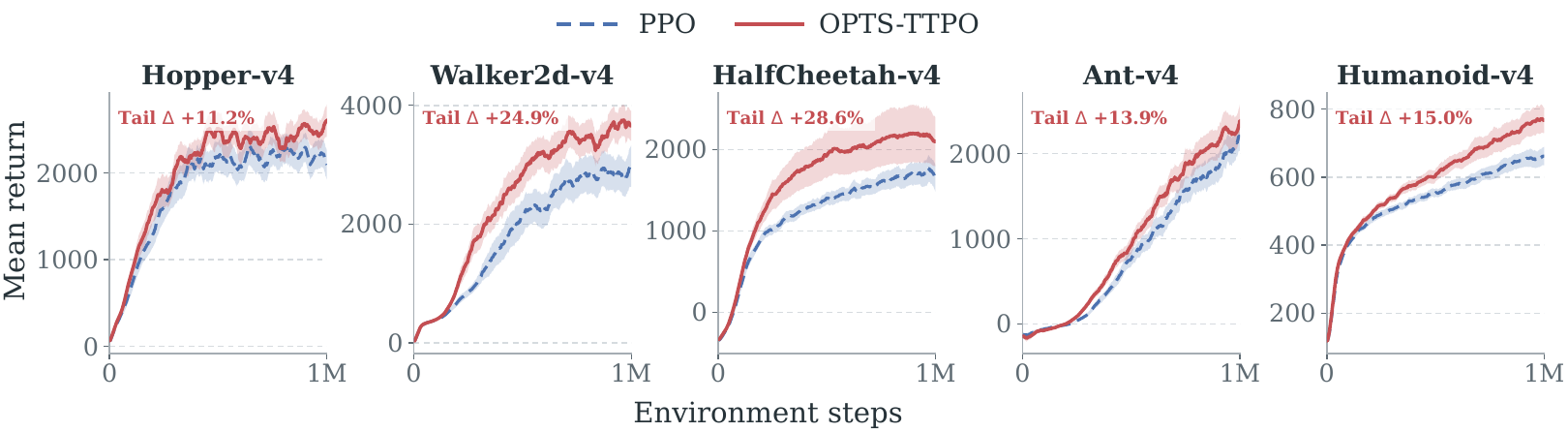}
\caption{MuJoCo learning curves; tail $\Delta$ is the final-100-point relative return improvement over PPO.}
\label{fig:mujoco-curves}
\end{figure}

\noindent\textbf{Continuous Control: MuJoCo.}\label{sec:mujoco} We select the hyperparameters on Hopper-v4 and Humanoid-v4, the two tasks with the smallest and largest action-space dimensions in our five-task MuJoCo suite~\citep{ref_mujoco}, respectively. We then fix $(\xi,S_{\max})=(0.6,1)$ for the five-task evaluation and train PPO and OPTS-TTPO for one million environment steps over ten random seeds (Figure~\ref{fig:mujoco-curves}).

OPTS-TTPO achieves a higher tail mean on all five tasks, with gains from $11.2\%$ on \texttt{Hopper-v4} to $28.6\%$ on \texttt{HalfCheetah-v4}, and a higher full-training mean on every task.

\Needspace{13\baselineskip}
\begin{wraptable}{r}{0.64\textwidth}
\vspace{-0.8\baselineskip}
\centering
\caption{\textbf{MuJoCo component ablations.} Tail-return change relative to PPO (\%), averaged over ten seeds at $(\xi,S_{\max})=(0.6,1)$. Tail return averages the final 100 logged points; Mean weights tasks equally.}
\label{tab:e4-fixed}
\small
\setlength{\tabcolsep}{2pt}
\resizebox{0.64\textwidth}{!}{%
\begin{tabular}{@{}lcccccc@{}}
\toprule
Variant & Hopper & Walker2d & HalfCheetah & Ant & Humanoid & Mean \\
\midrule
Random + mean & +10.5 & +5.6 & -3.2 & \textbf{+15.0} & -1.3 & +5.3 \\
Guided + mean & +8.2 & +16.7 & +9.5 & -0.6 & +6.3 & +8.0 \\
Guided + max, unweighted & +7.5 & +6.8 & +26.2 & +8.7 & +8.2 & +11.5 \\
Guided + max (full) & \textbf{+11.2} & \textbf{+24.9} & \textbf{+28.6} & +13.9 & \textbf{+15.0} & \textbf{+18.7} \\
\bottomrule
\end{tabular}
}
\end{wraptable}

\noindent\textbf{Component ablations.} At the fixed $(\xi,S_{\max})=(0.6,1)$, the complete method achieves the largest task-averaged tail improvement over PPO (Table~\ref{tab:e4-fixed}). Random + mean tests branching without guidance, Guided + mean adds adaptive selection, Guided + max adds max backup, and the unweighted row removes branch weights. Individual component gains vary by task.

\noindent\textbf{Stochastic Discrete Control: Atari-57.}\label{sec:atari} We reuse the same $(\xi,S_{\max})=(0.6,1)$ for Atari. Under the Atari-57 evaluation benchmark~\citep{ref_atari} with sticky actions~\citep{ref_sticky}, we train PPO and OPTS-TTPO for ten million environment steps over three random seeds. Because reward scales differ across games, we transform each run's return as $(R-R_{\mathrm{random}})/(R_{\mathrm{human}}-R_{\mathrm{random}})$ and report the interquartile mean (IQM), the 25\% trimmed mean over all 57 games and three seeds~\citep{ref_rliable}. Table~\ref{tab:atari-win-counts} pairs this magnitude-sensitive summary with task-level win counts under the full-training mean return and the last-100-log mean return; complete curves are in Appendix~\ref{app:atari-curves}.

\begin{wraptable}{r}{0.62\textwidth}
\vspace{-0.8\baselineskip}
\centering
\caption{Human-normalized IQM and task-level win counts on Atari-57. Each win compares the two methods after averaging the corresponding raw-return summary over three seeds for that game.}
\label{tab:atari-win-counts}
\small
\setlength{\tabcolsep}{2pt}
\resizebox{0.60\textwidth}{!}{%
\begin{tabular}{@{}lccccc@{}}
\toprule
& \multicolumn{2}{c}{Human-normalized IQM} & \multicolumn{3}{c}{Task wins} \\
\cmidrule(lr){2-3}\cmidrule(lr){4-6}
Metric & PPO & OPTS-TTPO & PPO & OPTS-TTPO & Tie \\
\midrule
Full-training mean return & 0.247 & \textbf{0.255} & 26 & \textbf{31} & 0 \\
Last-100-log mean return & 0.357 & \textbf{0.374} & 22 & \textbf{34} & 1 \\
\bottomrule
\end{tabular}
}
\end{wraptable}

At the task level, OPTS-TTPO posts a 31--26--0 win--loss--tie record against PPO under the full-training mean-return metric and a 34--22--1 record under the last-100-log mean-return metric. Its human-normalized IQM is also higher under both summaries, increasing from $0.247$ to $0.255$ over the full training run and from $0.357$ to $0.374$ over the last 100 logged points. These game-level records indicate that the gains depend on the task and its reward structure rather than holding uniformly across Atari-57. Appendix~\ref{app:atari-backup-ablation} compares mean and max backup at the same fixed hyperparameters.

\begin{table*}[!t]
\centering
\caption{LLM performance at the synchronized step-400 checkpoint, from 32 independent responses per prompt. Macro Average weights the six benchmarks equally; Micro Average pools their 500, 272, 40, 30, 30, and 30 problems, respectively (902 in total). Column bests are bold.}
\label{tab:llm-train-step400}
\scriptsize
\setlength{\tabcolsep}{4pt}
\renewcommand{\arraystretch}{1.12}
\resizebox{\textwidth}{!}{%
\begin{tabular}{@{}c@{\hspace{5pt}{\color{black!30}\vrule width 0.35pt}\hspace{4pt}}lcccccccccccccccc}
\toprule
\multicolumn{2}{c}{\multirow[c]{2}{*}{\textbf{Methods}}} & \multicolumn{2}{c}{\textbf{MATH500}} & \multicolumn{2}{c}{\textbf{MinervaMath}} & \multicolumn{2}{c}{\textbf{AMC23}} & \multicolumn{2}{c}{\textbf{AIME24}} & \multicolumn{2}{c}{\textbf{AIME25}} & \multicolumn{2}{c}{\textbf{AIME26}} & \multicolumn{2}{c}{\textbf{Macro Average}} & \multicolumn{2}{c}{\textbf{Micro Average}} \\
\cmidrule(lr){3-4}\cmidrule(lr){5-6}\cmidrule(lr){7-8}\cmidrule(lr){9-10}\cmidrule(lr){11-12}\cmidrule(lr){13-14}\cmidrule(lr){15-16}\cmidrule(lr){17-18}
\multicolumn{2}{c}{} & avg@32 & pass@32 & avg@32 & pass@32 & avg@32 & pass@32 & avg@32 & pass@32 & avg@32 & pass@32 & avg@32 & pass@32 & avg@32 & pass@32 & avg@32 & pass@32 \\
\midrule
\multicolumn{2}{@{}l}{\textbf{Qwen3-1.7B-Base}} \\
& PPO & 0.6973 & 0.9080 & \textbf{0.2986} & \textbf{0.5515} & 0.4289 & 0.8500 & 0.0823 & 0.3333 & 0.0469 & 0.3333 & 0.0417 & \textbf{0.2667} & 0.2659 & 0.5405 & 0.5013 & 0.7384 \\
& DAPO & 0.6935 & 0.9100 & 0.2878 & 0.5257 & 0.4133 & \textbf{0.8750} & 0.0854 & 0.3667 & 0.0490 & 0.3000 & 0.0385 & \textbf{0.2667} & 0.2612 & 0.5407 & 0.4953 & 0.7328 \\
& REINFORCE++ & 0.6877 & 0.9040 & 0.2911 & 0.5184 & 0.4016 & 0.8250 & 0.0823 & 0.3000 & \textbf{0.0521} & \textbf{0.3667} & 0.0354 & \textbf{0.2667} & 0.2584 & 0.5301 & 0.4924 & 0.7251 \\
& \textbf{OPTS-TTPO} & \textbf{0.7114} & \textbf{0.9220} & 0.2920 & 0.5404 & \textbf{0.4328} & 0.8000 & \textbf{0.1063} & \textbf{0.4333} & 0.0479 & 0.3000 & \textbf{0.0521} & \textbf{0.2667} & \textbf{0.2738} & \textbf{0.5437} & \textbf{0.5085} & \textbf{0.7428} \\
\midrule
\multicolumn{2}{@{}l}{\textbf{Qwen3-1.7B}} \\
& PPO & 0.7994 & 0.9480 & 0.3427 & 0.5257 & 0.5938 & 0.9250 & 0.2000 & \textbf{0.4667} & 0.1708 & \textbf{0.4000} & 0.1083 & 0.3333 & 0.3692 & 0.5998 & 0.5887 & 0.7650 \\
& DAPO & 0.8061 & 0.9280 & 0.3448 & 0.5221 & 0.5555 & 0.8750 & 0.1854 & 0.4000 & 0.1708 & 0.3000 & 0.1042 & 0.3000 & 0.3611 & 0.5542 & 0.5908 & 0.7439 \\
& REINFORCE++ & 0.7919 & 0.9400 & 0.3455 & 0.5368 & 0.5367 & 0.8500 & 0.1885 & 0.3667 & 0.1500 & 0.3667 & 0.0948 & 0.3333 & 0.3512 & 0.5656 & 0.5813 & 0.7561 \\
& \textbf{OPTS-TTPO} & \textbf{0.8271} & \textbf{0.9600} & \textbf{0.3532} & \textbf{0.5809} & \textbf{0.6281} & \textbf{0.9500} & \textbf{0.2354} & 0.4333 & \textbf{0.2031} & 0.3667 & \textbf{0.1333} & \textbf{0.3667} & \textbf{0.3967} & \textbf{0.6096} & \textbf{0.6119} & \textbf{0.7882} \\
\midrule
\multicolumn{2}{@{}l}{\textbf{Qwen3-8B-Base}} \\
& PPO & 0.8305 & 0.9360 & 0.4169 & 0.5699 & \textbf{0.6492} & 0.8500 & 0.1948 & 0.5000 & 0.1615 & 0.3000 & 0.1302 & 0.3333 & 0.3972 & 0.5815 & 0.6311 & 0.7661 \\
& DAPO & 0.8076 & 0.9460 & 0.3804 & 0.5368 & 0.5984 & 0.8250 & 0.1896 & 0.5000 & 0.1573 & \textbf{0.3667} & 0.1385 & 0.3000 & 0.3786 & 0.5791 & 0.6051 & 0.7616 \\
& REINFORCE++ & 0.8250 & 0.9360 & 0.4164 & 0.5478 & 0.6156 & 0.8750 & 0.1802 & \textbf{0.6000} & 0.1500 & 0.3333 & 0.1333 & 0.3333 & 0.3868 & 0.6042 & 0.6256 & 0.7650 \\
& \textbf{OPTS-TTPO} & \textbf{0.8514} & \textbf{0.9540} & \textbf{0.4241} & \textbf{0.5993} & 0.6055 & \textbf{0.9250} & \textbf{0.2031} & 0.4667 & \textbf{0.1813} & 0.3333 & \textbf{0.1458} & \textbf{0.3667} & \textbf{0.4019} & \textbf{0.6075} & \textbf{0.6443} & \textbf{0.7894} \\
\midrule
\multicolumn{2}{@{}l}{\textbf{Qwen3-8B}} \\
& PPO & 0.8799 & 0.9700 & 0.4323 & 0.5588 & 0.7094 & 0.9250 & \textbf{0.3531} & 0.5667 & 0.2615 & 0.5333 & 0.2594 & \textbf{0.5333} & 0.4826 & 0.6812 & 0.6786 & 0.8016 \\
& DAPO & \textbf{0.8859} & \textbf{0.9740} & 0.4493 & \textbf{0.5735} & \textbf{0.7633} & 0.9250 & 0.3208 & 0.6000 & 0.2656 & 0.4000 & \textbf{0.2667} & \textbf{0.5333} & 0.4919 & 0.6676 & \textbf{0.6888} & 0.8049 \\
& REINFORCE++ & 0.8749 & 0.9680 & 0.4323 & 0.5551 & 0.7164 & 0.9250 & 0.2938 & 0.5333 & 0.2406 & 0.4000 & 0.2073 & 0.5000 & 0.4609 & 0.6469 & 0.6718 & 0.7927 \\
& \textbf{OPTS-TTPO} & 0.8825 & \textbf{0.9740} & \textbf{0.4519} & 0.5699 & 0.7586 & \textbf{0.9500} & 0.3365 & \textbf{0.6667} & \textbf{0.2781} & \textbf{0.5667} & 0.2594 & \textbf{0.5333} & \textbf{0.4945} & \textbf{0.7101} & 0.6882 & \textbf{0.8126} \\
\bottomrule
\end{tabular}
}
\end{table*}

\noindent\textbf{Mathematical Reasoning: LLM RLVR.}\label{sec:llm-train-search} In LLM RLVR~\citep{ref_r1}, we compare OPTS-TTPO with PPO~\citep{ref_ppo} on four Qwen3 models~\citep{ref_qwen3}, using VeRL~\citep{ref_verl,ref_ray} and 4,096 rollouts per update. OPTS-TTPO uses $S_{\max}=3$ for all four models. DAPO~\citep{ref_dapo} and REINFORCE++~\citep{ref_reinforcepp} are external references at the same rollout budget; DAPO uses GRPO's group-relative objective~\citep{ref_grpo}. All methods run for 400 steps. Table~\ref{tab:llm-train-step400} reports final-checkpoint \texttt{avg@32} and \texttt{pass@32} on six benchmarks, including MATH500~\citep{ref_math500} and AIME25~\citep{ref_aime25}.

OPTS-TTPO improves both macro- and micro-averaged \texttt{avg@32} and \texttt{pass@32} over PPO across all four Qwen3 models, covering both model sizes and both Base and post-trained variants. Among the four methods, it achieves the highest macro averages for both metrics on every model. The micro averages show the same ranking except for \texttt{avg@32} on Qwen3-8B, where DAPO scores $0.6888$ and OPTS-TTPO scores $0.6882$.

\takeaway{4}{Across matched budgets, OPTS-TTPO improves tail returns on all five MuJoCo tasks, posts a 34--22--1 win--loss--tie record against PPO on Atari-57 under the last-100-log mean-return metric, and improves micro-averaged \texttt{avg@32} and \texttt{pass@32} over PPO across all four Qwen3 models.}

%% file: sections/5_Conclusion.tex
\section{Conclusion}

We study how tree search can improve finite-sample policy-gradient learning by increasing high-return trajectory coverage. We first make tree trajectories usable for learning: branch aggregation yields TTPG, TreeGAE, and TTPO, and the branch-aggregation experiment shows that TTPG stays near the unbranched reference as suffixes are added, whereas naive aggregation accumulates bias. We then use performance-difference-guided OPTS to allocate a finite search budget. Exact-value experiments show improvements in both expected $\lambda$-return and true return with every adjacent budget increase, while learned-critic experiments show the same aggregate trend. At matched test-time budgets, reward-guided OPTS improves correct-answer coverage and value-guided OPTS improves majority-vote accuracy over independent sampling. At matched branch counts, coverage--bias diagnostics show that OPTS + TTPG trades a modest bias increase for greater coverage relative to Fixed-branch + TTPG; max backup adds additional prefix credit, while mean backup averages sampled suffixes in stochastic environments. Across domains, OPTS-TTPO improves tail returns on all five MuJoCo tasks, posts a 34--22--1 win--loss--tie record against PPO on Atari-57 under the last-100-log mean-return metric, and improves micro-averaged \texttt{avg@32} and \texttt{pass@32} over PPO across all four Qwen3 models at matched budgets. However, the theory bounds the bias from posterior-dependent expansion, rather than eliminating it, while the monotone-improvement guarantee is limited to deterministic dynamics with exact values. Future work includes variance reduction under prefix-measurable sampling and self-distillation from successful OPTS trajectories.

%% file: sections/Appendix.tex
\numberwithin{equation}{section}
\numberwithin{figure}{section}
\numberwithin{table}{section}
\input{sections/Appendix_methods.tex}
\input{sections/Appendix_experiments.tex}

%% file: sections/Appendix_methods.tex
\numberwithin{equation}{section}


\section{Tree Trajectory Policy Optimization}
\label{app:ttpo-theory}

This section derives Tree Trajectory Policy Optimization (TTPO) from the standard chain-trajectory policy gradient. We first define a general branch aggregation rule for chain statistics, and then instantiate its direct and recursive forms as the Tree Trajectory Policy Gradient (TTPG) and Tree-based Generalized Advantage Estimation (TreeGAE).

\subsection{Standard Chain Policy Gradient and Generalized Advantage Estimation}

Consider a finite-horizon discounted Markov decision process
\[
\mathcal{M}=(\mathcal{S},\mathcal{A},P,r,\rho_0,\gamma,n)
\]
and a differentiable stochastic policy $\pi_\theta(a\mid s)$. The time index is included in the state. A chain trajectory is written as $\tau=(x_0,\ldots,x_{n-1})$, where $x_t=(s_t,a_t,r_t,s_{t+1})$, $s_0\sim\rho_0$, $a_t\sim\pi_\theta(\cdot\mid s_t)$, and $(r_t,s_{t+1})\sim P(\cdot\mid s_t,a_t)$. We set $V^\pi(s_n)=0$ and pad an earlier terminal state by zero-reward absorbing transitions when necessary. The discounted objective and suffix return are
\begin{equation}
\label{eq:appendix-chain-objective}
J(\theta)
:=
\mathbb{E}_{\tau\sim\pi_\theta}
\left[\sum_{t=0}^{n-1}\gamma^t r_t\right],
\qquad
G_t
:=
\sum_{k=t}^{n-1}\gamma^{k-t}r_k .
\end{equation}
Let $Q^\pi(s_t,a_t):=\mathbb{E}[G_t\mid s_t,a_t]$, $V^\pi(s_t):=\mathbb{E}_{a\sim\pi(\cdot\mid s_t)}[Q^\pi(s_t,a)]$, and $A^\pi(s_t,a_t):=Q^\pi(s_t,a_t)-V^\pi(s_t)$. The likelihood-ratio identity gives
\begin{align}
\nabla_\theta J(\theta)
&=
\mathbb{E}_{\tau\sim\pi_\theta}
\left[
\sum_{t=0}^{n-1}
\nabla_\theta\log\pi_\theta(a_t\mid s_t)
\sum_{k=t}^{n-1}\gamma^k r_k
\right]
\nonumber\\
&=
\mathbb{E}_{\tau\sim\pi_\theta}
\left[
\sum_{t=0}^{n-1}
\gamma^t Q^{\pi_\theta}(s_t,a_t)
\nabla_\theta\log\pi_\theta(a_t\mid s_t)
\right].
\label{eq:appendix-chain-pg-q}
\end{align}
For any state-dependent baseline $b(s_t)$,
$\mathbb{E}_{a_t\sim\pi_\theta(\cdot\mid s_t)}[b(s_t)\nabla_\theta\log\pi_\theta(a_t\mid s_t)]=0$. Taking $b=V^{\pi_\theta}$ yields the standard chain policy gradient~\citep{sutton1999policy}:
\begin{equation}
\label{eq:appendix-chain-policy-gradient}
\nabla_\theta J(\theta)
=
\mathbb{E}_{\tau\sim\pi_\theta}
\left[
\sum_{t=0}^{n-1}
\gamma^t A^{\pi_\theta}(s_t,a_t)
\nabla_\theta\log\pi_\theta(a_t\mid s_t)
\right].
\end{equation}

For any value function $V$, define the temporal-difference residual
\begin{equation}
\label{eq:appendix-chain-td}
\delta_t^V
:=
r_t+\gamma V(s_{t+1})-V(s_t).
\end{equation}
Generalized Advantage Estimation (GAE)~\citep{ref_gae} is the discounted chain sum of TD residuals,
\begin{equation}
\label{eq:appendix-chain-gae}
\widehat A_t^{\mathrm{GAE}(V)}
:=
\sum_{k=t}^{n-1}(\gamma\lambda)^{k-t}\delta_k^V
=
\delta_t^V+\gamma\lambda\widehat A_{t+1}^{\mathrm{GAE}(V)},
\qquad
\widehat A_n^{\mathrm{GAE}(V)}=0.
\end{equation}
When $V=V^{\pi_\theta}$, the first TD residual satisfies
$\mathbb{E}[\delta_t^{V^{\pi_\theta}}\mid s_t,a_t]=A^{\pi_\theta}(s_t,a_t)$. For every $k>t$, on-policy action sampling and the Markov property give
\begin{equation}
\mathbb{E}
\left[
\delta_k^{V^{\pi_\theta}}
\mid s_t,a_t
\right]
=
\mathbb{E}
\left[
\mathbb{E}
\left[
\delta_k^{V^{\pi_\theta}}
\mid s_k
\right]
\middle|
s_t,a_t
\right]
=0.
\end{equation}
Consequently,
\begin{equation}
\label{eq:appendix-chain-gae-unbiased}
\mathbb{E}
\left[
\widehat A_t^{\mathrm{GAE}(V^{\pi_\theta})}
\mid s_t,a_t
\right]
=
A^{\pi_\theta}(s_t,a_t).
\end{equation}

\subsection{On-Policy Tree Trajectories}

An on-policy tree trajectory $\mathcal{T}=(s_o,\mathcal{X}(\mathcal{T}))$ consists of a root state $s_o\sim\rho_0$ and a finite indexed set $\mathcal{X}(\mathcal{T})$ of transition occurrences. The index $o$ identifies the root-state occurrence. Starting from $s_o$, the construction recursively samples outgoing transitions and extends their successor occurrences to the horizon; suffixes sharing a state occurrence share the prefix ending there. A transition from a parent state occurrence $s_p$ to a child state occurrence $s_c$ is written as $(s_p,a_c,r_c,s_c)$. The transition set satisfies the following conditions. First, rooted connectivity requires every transition to have a unique path from $s_o$: its source is either $s_o$ or the successor-state occurrence of exactly one preceding transition. Second, every transition is generated by the current policy and the environment,
\begin{equation}
\label{eq:appendix-on-policy-node-generation}
a_c\sim\pi_\theta(\cdot\mid s_p),
\qquad
(r_c,s_c)\sim P(\cdot\mid s_p,a_c).
\end{equation}
When the parent--child indices are not needed, we index transition occurrences by $x$ and denote their source state, action, reward, and successor state by $s_x,a_x,r_x,s_{x^+}$, respectively. The successor state $s_{x^+}$ is also the source state of each child transition $x^+$. The depth of $x$ is $d(x)\in\{0,\ldots,n-1\}$, and $x\in\mathcal{T}$ means $x\in\mathcal{X}(\mathcal{T})$.

Here, \emph{on-policy} refers to the conditional generation in Equation~\ref{eq:appendix-on-policy-node-generation}. Because every new suffix is sampled from the current policy, tree search introduces no separate behavior policy and requires no additional action-distribution importance correction at data collection. This local property does not imply that the random tree has the same state-visitation distribution as a policy chain. Selecting a previously visited state for rebranching is a decision made by an external algorithm rather than an action generated by the policy; states preferred by the branching rule therefore receive more suffix samples in the tree. The Branch Aggregation Lemma below gives the conditions under which a branch-weighted additive statistic on such a tree recovers the expectation of its chain-trajectory counterpart.

\subsection{Branch Aggregation Lemma}
\label{app:branch-aggregation}

The policy gradient and GAE are expectations of discounted sums along chain trajectories, and a tree changes how often each part of a chain is observed. Section~\ref{sec:branch-aggregation-lemma} illustrates this with three suffixes sampled at one depth of a chain: naive summation counts the suffix three times relative to the prefix, and averaging the three suffixes restores the chain proportions. The lemma generalizes this correction to arbitrary trees: it gives the weights under which tree sums recover the chain expectation, and states the conditions on how branches may be chosen for the correction to hold.

Let $h$ be any integrable scalar- or vector-valued function of a transition and let $\eta$ be a scalar. The chain statistic of interest is
\begin{equation}
\label{eq:appendix-chain-statistic}
H_\eta(\tau)
:=
\sum_{t=0}^{n-1}\eta^t h(s_t,a_t,r_t,s_{t+1}).
\end{equation}
Its on-policy target is $\mathbb{E}_{\tau\sim\pi_\theta}[H_\eta(\tau)]$.

Let $\mathcal F_p$ denote the sigma-field of all information available upon reaching $s_p$, before sampling its outgoing transitions. Write $\mathcal C(p)$ for its child-index set and $\alpha_{p,c}$ for a local branch weight. In the generic transition notation $x$, the corresponding information before sampling $a_x$ is denoted by $\mathcal F_x$.

\noindent\textbf{Conditions for recovering the chain expectation.} \begin{enumerate}
\item \textbf{Conditional on-policy sampling.} Given $\mathcal F_p$, each outgoing transition $c\in\mathcal C(p)$ is sampled according to Equation~\ref{eq:appendix-on-policy-node-generation}.
\item \textbf{Measurable, normalized branch weights.} The finite local-weight family $(\alpha_{p,c})_{c\in\mathcal C(p)}$ is $\mathcal F_p$-measurable and satisfies
\begin{equation}
\label{eq:appendix-local-branch-weights}
\alpha_{p,c}\ge0,
\qquad
\sum_{c\in\mathcal{C}(p)}\alpha_{p,c}=1.
\end{equation}
\end{enumerate}

The child set and its local weights may be random and prefix-dependent, but they must be $\mathcal F_p$-measurable, that is, determined before the corresponding outgoing transitions are sampled. Conditional on-policy sampling alone is therefore insufficient for Equation~\ref{eq:appendix-branch-aggregation}.

Additional on-policy suffixes may be sampled from selected state occurrences and share their existing prefixes. Further branching may occur along the resulting suffixes, while unbranched segments retain the ordinary chain structure. The local contribution of a transition is still $h(c):=h(s_p,a_c,r_c,s_c)$, and a shared prefix appears once in the tree. Along a branch, the available information satisfies $\mathcal F_p\subseteq\mathcal F_c$.

The global branch weight is computed recursively as
\begin{equation}
\label{eq:appendix-global-branch-weight}
W(o):=1,
\qquad
W(c):=W(p)\alpha_{p,c},
\quad c\in\mathcal C(p).
\end{equation}
Thus $W(c)$ is the product of the local weights along the unique path from $s_o$ to $c$. We next define the branch-weighted tree statistic by direct summation over all transitions and by recursive aggregation from the leaves to the root.

\noindent\textbf{Direct aggregation.} The direct branch-weighted tree sum is
\begin{equation}
\label{eq:appendix-direct-tree-aggregation}
\widehat H_{\eta,\mathrm{dir}}(\mathcal{T})
:=
\sum_{c\in\mathcal{X}(\mathcal{T})}W(c)\eta^{d(c)}h(c).
\end{equation}

\noindent\textbf{Recursive aggregation.} Let $\widehat R_\eta(s_p)$ denote the aggregated suffix beginning at the state occurrence $s_p$. Starting from the leaves, define
\begin{equation}
\label{eq:appendix-recursive-tree-aggregation}
\widehat R_\eta(s_p)
:=
\sum_{c\in\mathcal{C}(p)}
\alpha_{p,c}
\left[h(c)+\eta\widehat R_\eta(s_c)\right],
\end{equation}
where $\widehat R_\eta(s_c)=0$ when $d(c)=n-1$. At the root, define
\begin{equation}
\label{eq:appendix-recursive-tree-root}
\widehat H_{\eta,\mathrm{rec}}(\mathcal{T})
:=
\widehat R_\eta(s_o).
\end{equation}

\noindent\textbf{Lemma 1 (Branch aggregation).} For every realized tree, the two aggregation forms satisfy
\begin{equation}
\label{eq:appendix-direct-recursive-equivalence}
\widehat H_{\eta,\mathrm{dir}}(\mathcal{T})
=
\widehat H_{\eta,\mathrm{rec}}(\mathcal{T}).
\end{equation}
Under the recovery conditions above, both forms recover the expectation of the original chain statistic:
\begin{equation}
\label{eq:appendix-branch-aggregation}
\mathbb{E}_{\mathcal{T}}
\left[
\widehat H_{\eta,\mathrm{dir}}(\mathcal{T})
\right]
=
\mathbb{E}_{\mathcal{T}}
\left[
\widehat H_{\eta,\mathrm{rec}}(\mathcal{T})
\right]
=
\mathbb{E}_{\tau\sim\pi_\theta}
\left[
H_\eta(\tau)
\right].
\end{equation}

\noindent\textbf{Proof.} \textbf{Part I: Direct--recursive equivalence.}

\emph{Root layer.}
Equations~\ref{eq:appendix-global-branch-weight} and~\ref{eq:appendix-recursive-tree-aggregation} give
\begin{align}
\widehat R_\eta(s_o)
&=
\sum_{c\in\mathcal C(o)}
\alpha_{o,c}
\left[h(c)+\eta\widehat R_\eta(s_c)\right]
\nonumber\\
&=
\sum_{d(c)=0}W(c)h(c)
+
\eta\sum_{d(c)=0}W(c)\widehat R_\eta(s_c).
\label{eq:appendix-root-layer-expansion}
\end{align}

\emph{One-layer expansion.}
For every $k\in\{0,\ldots,n-2\}$, expanding the depth-$k$ remainder once gives
\begin{align}
&\eta^{k+1}
\sum_{d(p)=k}
W(p)\widehat R_\eta(s_p)
\nonumber\\
={}&
\eta^{k+1}
\sum_{d(p)=k}
\sum_{c\in\mathcal C(p)}
W(p)\alpha_{p,c}
\left[h(c)+\eta\widehat R_\eta(s_c)\right]
\nonumber\\
={}&
\eta^{k+1}
\sum_{d(c)=k+1}W(c)h(c)
+
\eta^{k+2}
\sum_{d(c)=k+1}W(c)\widehat R_\eta(s_c).
\label{eq:appendix-one-layer-expansion}
\end{align}

\emph{Full expansion.}
Applying Equation~\ref{eq:appendix-one-layer-expansion} successively to Equation~\ref{eq:appendix-root-layer-expansion}, and using $\widehat R_\eta(s_c)=0$ at $d(c)=n-1$, yields
\begin{align}
\widehat H_{\eta,\mathrm{rec}}(\mathcal T)
&=
\widehat R_\eta(s_o)
=
\sum_{k=0}^{n-1}
\sum_{d(c)=k}
W(c)\eta^k h(c)
\nonumber\\
&=
\sum_{c\in\mathcal X(\mathcal T)}
W(c)\eta^{d(c)}h(c)
=
\widehat H_{\eta,\mathrm{dir}}(\mathcal T).
\label{eq:appendix-expanded-direct-recursive-equivalence}
\end{align}

\textbf{Part II: Recovery of the chain expectation.}

\emph{Chain suffix target.}
Let $\tau_p$ be an on-policy chain suffix from $s_p$ through the remaining horizon, and let $H_\eta(\tau_p)$ denote its remaining chain sum with the local discount starting from $\eta^0$; at the terminal state this sum is zero. Define
\begin{equation}
\mu_\eta(p)
:=
\mathbb E_{\tau_p\sim\pi_\theta}
\left[
H_\eta(\tau_p)
\mid\mathcal F_p
\right].
\label{eq:appendix-chain-suffix-mean}
\end{equation}

\emph{One-step chain recursion.}
Equation~\ref{eq:appendix-on-policy-node-generation} and the tower property give, for every $c\in\mathcal C(p)$,
\begin{equation}
\mathbb E_{\mathcal T}
\left[
h(c)+\eta\mu_\eta(c)
\mid\mathcal F_p
\right]
=
\mu_\eta(p).
\label{eq:appendix-on-policy-suffix-recursion}
\end{equation}

\emph{Base case.}
At $d(c)=n-1$,
\begin{equation}
\mathbb E_{\mathcal T}
\left[
\widehat R_\eta(s_c)
\mid\mathcal F_c
\right]
=0
=\mu_\eta(c).
\label{eq:appendix-aggregation-induction-base}
\end{equation}

\emph{Induction step.}
Assuming the same identity at the children of $p$, Equations~\ref{eq:appendix-recursive-tree-aggregation} and~\ref{eq:appendix-on-policy-suffix-recursion} give
\begin{align}
\mathbb E_{\mathcal T}
\left[
\widehat R_\eta(s_p)
\mid\mathcal F_p
\right]
&=
\sum_{c\in\mathcal C(p)}
\alpha_{p,c}
\mathbb E_{\mathcal T}
\left[
h(c)
+
\eta
\mathbb E_{\mathcal T}
\left[
\widehat R_\eta(s_c)
\mid\mathcal F_c
\right]
\middle|\mathcal F_p
\right]
\nonumber\\
&=
\sum_{c\in\mathcal C(p)}
\alpha_{p,c}
\mathbb E_{\mathcal T}
\left[
h(c)+\eta\mu_\eta(c)
\mid\mathcal F_p
\right]
\nonumber\\
&=
\sum_{c\in\mathcal C(p)}
\alpha_{p,c}\mu_\eta(p)
=
\mu_\eta(p).
\label{eq:appendix-aggregation-induction-step}
\end{align}

\emph{Root expectation.}
Finally, $s_o\sim\rho_0$ and the tower property give
\begin{align}
\mathbb E_{\mathcal T}
\left[
\widehat H_{\eta,\mathrm{rec}}(\mathcal T)
\right]
&=
\mathbb E_{\mathcal T}
\left[
\mathbb E_{\mathcal T}
\left[
\widehat R_\eta(s_o)
\mid\mathcal F_o
\right]
\right]
\nonumber\\
&=
\mathbb E_{s_o\sim\rho_0}
\left[
\mu_\eta(o)
\right]
=
\mathbb E_{\tau\sim\pi_\theta}
\left[
H_\eta(\tau)
\right].
\label{eq:appendix-recursive-unbiasedness}
\end{align}
Together with Equation~\ref{eq:appendix-expanded-direct-recursive-equivalence}, this proves Equation~\ref{eq:appendix-branch-aggregation}.
\hfill$\square$

\subsection{Applying the Branch Aggregation Lemma to Tree Trajectories}

The tree generator in this subsection satisfies the chain-expectation recovery conditions in Appendix~\ref{app:branch-aggregation}; in particular, its transitions are conditionally sampled from $\pi_\theta$ and its branch weights are fixed from prefix information before the corresponding outgoing transitions are sampled.

\noindent\textbf{Tree Trajectory Policy Gradient (TTPG).} The branching decisions and weights are held fixed under differentiation. Set $\eta=\gamma$ and
\[
h(x)
=
A^{\pi_\theta}(s_x,a_x)
\nabla_\theta\log\pi_\theta(a_x\mid s_x)
\]
in the direct aggregation formula. Applying Lemma~1 and Equation~\ref{eq:appendix-chain-policy-gradient} gives
\begin{equation}
\label{eq:appendix-weighted-tree-policy-gradient}
\nabla_\theta J(\theta)
=
\mathbb{E}_{\mathcal{T}}
\left[
\sum_{x\in\mathcal{T}}
W(x)\gamma^{d(x)}
A^{\pi_\theta}(s_x,a_x)
\nabla_\theta\log\pi_\theta(a_x\mid s_x)
\right]
.
\end{equation}
Thus TTPG is the direct-summation instance of branch aggregation.

\noindent\textbf{Tree-based Generalized Advantage Estimation (TreeGAE).} For a transition $x$, define
\begin{equation}
\label{eq:appendix-tree-td}
\delta_x^V
:=
r_x+\gamma V(s_{x^+})-V(s_x).
\end{equation}
Set $\eta=\gamma\lambda$ and $h(x)=\delta_x^V$ in recursive aggregation. The resulting TreeGAE recursion is
\begin{equation}
\label{eq:appendix-tree-gae}
\widehat A_x^{\mathrm{TreeGAE}(V)}
:=
\delta_x^V
+
\gamma\lambda
\sum_{x^+\in\mathcal{C}(x)}
\alpha_{x,x^+}
\widehat A_{x^+}^{\mathrm{TreeGAE}(V)},
\end{equation}
with the recursive term set to zero beyond depth $n-1$.

Applying Lemma~1 to the transition subtree beginning at $x$ shows that TreeGAE and chain GAE have the same conditional suffix expectation under the same value function. For $V=V^{\pi_\theta}$, Equation~\ref{eq:appendix-chain-gae-unbiased} and the Markov property give
\begin{equation}
\label{eq:appendix-tree-gae-unbiased}
\mathbb{E}
\left[
\widehat A_x^{\mathrm{TreeGAE}(V^{\pi_\theta})}
\mid \mathcal{F}_x,a_x
\right]
=
\mathbb{E}
\left[
\widehat A_t^{\mathrm{GAE}(V^{\pi_\theta})}
\mid s_t,a_t
\right]
=
A^{\pi_\theta}(s_x,a_x).
\end{equation}
By Equation~\ref{eq:appendix-tree-gae-unbiased} and the $\mathcal F_x$-measurability of $W(x)$,
\begin{align}
&\mathbb{E}
\left[
W(x)\gamma^{d(x)}
\widehat A_x^{\mathrm{TreeGAE}(V^{\pi_\theta})}
\nabla_\theta\log\pi_\theta(a_x\mid s_x)
\middle|
\mathcal{F}_x,a_x
\right]
\nonumber\\
={}&
W(x)\gamma^{d(x)}
A^{\pi_\theta}(s_x,a_x)
\nabla_\theta\log\pi_\theta(a_x\mid s_x).
\label{eq:appendix-tree-gae-score-conditioning}
\end{align}
Summing over the tree and applying the tower property yields
\begin{equation}
\label{eq:appendix-tree-gae-policy-gradient}
\mathbb{E}_{\mathcal{T}}
\left[
\sum_{x\in\mathcal{T}}
W(x)\gamma^{d(x)}
\widehat A_x^{\mathrm{TreeGAE}(V^{\pi_\theta})}
\nabla_\theta\log\pi_\theta(a_x\mid s_x)
\right]
=
\nabla_\theta J(\theta).
\end{equation}

\subsection{PPO-Style Tree Trajectory Policy Optimization}

In practice, as in PPO, we omit the outer state-visitation discount by setting $\gamma^{d(x)}=1$ in the actor and critic objectives and the normalization statistics below. The discount used in return estimation and TreeGAE remains unchanged.

Let $\pi_{\theta_{\mathrm{old}}}$ be the policy used to sample the actions in the on-policy tree and define the per-node probability ratio
\begin{equation}
\label{eq:appendix-ttpo-ratio}
\rho_x(\theta)
:=
\frac{\pi_\theta(a_x\mid s_x)}
{\pi_{\theta_{\mathrm{old}}}(a_x\mid s_x)}.
\end{equation}
For a minibatch $\mathcal B$ of sampled trees, let $N_{\mathcal B}:=\sum_{\mathcal T\in\mathcal B}|\mathcal X(\mathcal T)|$ be the number of transitions it contains, and define the branch-weighted empirical average by
\begin{equation}
\mathbb E_x[W(x)f(x)]
:=
\frac{1}{N_{\mathcal B}}
\sum_{\mathcal T\in\mathcal B}
\sum_{x\in\mathcal T}W(x)f(x).
\label{eq:appendix-transition-mean}
\end{equation}
Search changes how rollout suffixes share prefixes and therefore how many transition occurrences are stored in a minibatch. Following the per-transition averaging used in PPO implementations, we divide the branch-weighted actor and critic sums by $N_{\mathcal B}$, so that the raw number of stored transitions does not directly scale the update. This is a practical loss normalization rather than part of the chain-expectation identity in Lemma~1.
With the tree and its branch weights held fixed, the clipped TTPO actor objective extends the PPO surrogate~\citep{ref_ppo} to action-level gradient aggregation:
\begin{equation}
\mathcal{L}_{\pi}^{\mathrm{TTPO}}(\theta)
:=
\mathbb{E}_x
\left[
W(x)
\min\Big(
\rho_x(\theta)\widehat A_x,
\operatorname{clip}(\rho_x(\theta),1-\epsilon,1+\epsilon)
\widehat A_x
\Big)
\right],
\label{eq:appendix-ttpo-actor}
\end{equation}
where $\widehat A_x$ is fixed when optimizing $\theta$. At $\theta=\theta_{\mathrm{old}}$, $\rho_x(\theta_{\mathrm{old}})=1$, and the derivative of either clipped branch is
$\widehat A_x\nabla_\theta\log\pi_\theta(a_x\mid s_x)$. Hence
\begin{equation}
\label{eq:appendix-ttpo-first-order}
\left.
\nabla_\theta
\mathcal{L}_{\pi}^{\mathrm{TTPO}}(\theta)
\right|_{\theta=\theta_{\mathrm{old}}}
=
\mathbb{E}_x
\left[
W(x)
\widehat A_x
\nabla_{\theta_{\mathrm{old}}}\log\pi_{\theta_{\mathrm{old}}}(a_x\mid s_x)
\right].
\end{equation}
Thus maximizing $\mathcal{L}_{\pi}^{\mathrm{TTPO}}$ gives the PPO-style clipped counterpart of TTPG under the practical convention above.

For value estimation, let $\bar\phi$ denote the critic parameters used to construct the fixed target, define $\widehat R_x:=\widehat A_x+V_{\bar\phi}(s_x)$, and define the clipped value prediction
\begin{equation}
V_\phi^{\mathrm{clip}}(s_x)
:=
V_{\bar\phi}(s_x)
+
\operatorname{clip}
\left(
V_\phi(s_x)-V_{\bar\phi}(s_x),
-\epsilon_V,
\epsilon_V
\right).
\end{equation}
The corresponding branch-weighted value objective is
\begin{equation}
\label{eq:appendix-ttpo-critic}
\mathcal{L}_{V}^{\mathrm{TTPO}}(\phi)
:=
\mathbb{E}_x
\left[
W(x)
\max\left(
(V_\phi(s_x)-\widehat R_x)^2,
(V_\phi^{\mathrm{clip}}(s_x)-\widehat R_x)^2
\right)
\right].
\end{equation}
The actor objective is maximized and the value objective is minimized.

\noindent\textbf{Branch-weighted advantage normalization.}\label{app:advantage-normalization} Whitening serves a different purpose and therefore uses the total branch mass $Z_{\mathcal B}:=\sum_{\mathcal T\in\mathcal B}\sum_{x\in\mathcal T}W(x)$ to normalize the branch weights. Under the conditions of Lemma~1, the resulting weighted statistics are consistent for the corresponding moments of the chain-transition distribution; using $N_{\mathcal B}$ instead would leave the branch weights unnormalized and allow branch multiplicity to distort these moments. As in PPO, whitening with finite-batch statistics remains an approximation. Define the weighted advantage mean and degrees-of-freedom-corrected variance by
\begin{equation}
\label{eq:appendix-whitening-moments}
\begin{aligned}
\mu_A
&:=
\frac{1}{Z_{\mathcal{B}}}
\sum_{\mathcal{T}\in\mathcal{B}}
\sum_{x\in\mathcal{T}}
W(x)\widehat A_x,
\\
\sigma_A^2
&:=
\frac{
\sum_{\mathcal{T}\in\mathcal{B}}
\sum_{x\in\mathcal{T}}
W(x)
(\widehat A_x-\mu_A)^2
}{
Z_{\mathcal B}
-Z_{\mathcal B}^{-1}
\sum_{\mathcal T\in\mathcal B}\sum_{x\in\mathcal T}W(x)^2
}.
\end{aligned}
\end{equation}
The normalized advantage is
\begin{equation}
\label{eq:appendix-weighted-normalized-advantage}
\widetilde A_x
:=
\frac{\widehat A_x-\mu_A}
{\sqrt{\sigma_A^2+\varepsilon}}.
\end{equation}
When advantage normalization is enabled, $\widetilde A_x$ replaces $\widehat A_x$ in Equation~\ref{eq:appendix-ttpo-actor}.

\noindent\textbf{Scope and transition to adaptive search.} Unlike the tree generators covered above, OPTS, introduced in the next section, selects expansion states using outcomes already observed in the tree. Its resulting child sets and local weights therefore need not be $\mathcal F_p$-measurable and can be correlated with the suffix statistics they aggregate. Consequently, the conditional-expectation step in Equation~\ref{eq:appendix-aggregation-induction-step} does not directly extend to OPTS trees, and Lemma~1 alone does not establish chain-expectation recovery for the adaptive search distribution. Appendix~\ref{app:opts-ttpo-theory} separates this posterior selection effect from the additional max-backup prefix-credit term and bounds the resulting gradient bias.

\section{On-Policy Parallel Tree Search}
\label{app:opts-theory}

This section derives a performance-difference estimate for rebranching by accumulating policy-relative local action improvements, aligns the estimate with the domain's search-budget unit, and uses it to define On-Policy Parallel Tree Search (OPTS). This is the budget-allocation layer of the method.

\subsection{Performance-Difference Estimation}
\label{app:perf-diff-estimation}

Let $\pi$ be the policy used to resample a suffix, and let $\mu$ be an arbitrary behavior policy. Starting from $s_t$, $\mu$ and the environment generate $\tau=(s_t,a_t,r_t,\ldots,s_{n-1},a_{n-1},r_{n-1},s_n)\sim(\mu,P)$, where $s_n$ is terminal and $V^\pi(s_n)=0$.

\noindent\textbf{Performance-Difference Estimate.} We define the performance-difference estimate at $s_t$ on $\tau$ as
\begin{equation}
\label{eq:appendix-negative-suffix-advantage}
\Delta(s_t;\tau)
:=
-\sum_{k=t}^{n-1}\gamma^{k-t}A^\pi(s_k,a_k).
\end{equation}
Each term $-A^\pi(s_k,a_k)=V^\pi(s_k)-Q^\pi(s_k,a_k)$ is the local expected improvement from replacing $a_k$ by a fresh action from $\pi$ and following $\pi$ thereafter. The estimate has two properties:
\begin{enumerate}
\item \textbf{Expected performance improvement.}
The performance difference lemma~\citep{kakade2002approximately,schulman2015trust} gives
\begin{equation}
\label{eq:appendix-perf-diff}
\mathbb E_{\tau\sim(\mu,P)\mid s_t}
\!\left[\Delta(s_t;\tau)\right]
=
V^\pi(s_t)-V^\mu(s_t).
\end{equation}

Thus $\Delta(s_t;\tau)$ estimates the expected return improvement from switching from $\mu$ to $\pi$ at $s_t$.

In particular, when $\mu=\pi$, Equation~\ref{eq:appendix-perf-diff} becomes
\begin{equation}
\mathbb E_{\tau\sim(\pi,P)\mid s_t}
\!\left[\Delta(s_t;\tau)\right]
=0.
\label{eq:appendix-on-policy-perf-diff}
\end{equation}
Before observing the reference suffix, retaining it and independently resampling from the same policy have equal expected value.

\item \textbf{Improvement in deterministic environments.}
When the environment is deterministic, $Q^\pi(s_k,a_k)=r_k+\gamma V^\pi(s_{k+1})$, and Equation~\ref{eq:appendix-negative-suffix-advantage} telescopes to
\begin{equation}
\Delta(s_t;\tau)
=
V^\pi(s_t)
-
\sum_{k=t}^{n-1}\gamma^{k-t}r_k.
\label{eq:appendix-deterministic-perf-diff}
\end{equation}
The first term is the expected return of a new suffix sampled from $\pi$, and the second is the return of the observed reference suffix. Their difference is the expected improvement from rebranching at $s_t$.
\end{enumerate}

\noindent\textbf{Rebranching state selection.} The rebranching state is the admissible state on the observed path $\tau$ with the largest performance-difference estimate:
\begin{equation}
\label{eq:appendix-rebranch-state}
s^\star
\in
\arg\max_{s_t\in\tau}\Delta(s_t;\tau).
\end{equation}

\noindent\textbf{TreeGAE estimation.} In practice, TreeGAE advantages $\widehat A_{x_k}$ replace $A^\pi$, giving the TreeGAE-estimated performance difference:
\begin{equation}
\label{eq:appendix-estimated-perf-diff}
\widehat\Delta(s_t;\tau)
:=
-\sum_{k=t}^{n-1}
\gamma^{k-t}\widehat A_{x_k},
\qquad
s^\star
\in
\arg\max_{s_t\in\tau}
\widehat\Delta(s_t;\tau).
\end{equation}

\subsection{Budget Alignment and Length Penalty Across Domains}
\label{app:budget-alignment}

The search budget is measured in rollouts for LLMs and in transitions for Atari and MuJoCo. Accordingly, we use the unpenalized estimate for LLMs and a suffix-length-penalized estimate for Atari and MuJoCo:
\begin{equation}
\label{eq:appendix-length-penalized-estimate}
\widehat\Delta^{(\xi)}(s_k;\tau)
=
\frac{\widehat\Delta(s_k;\tau)}{(n-k)^\xi},
\qquad
\xi\in[0,1] .
\end{equation}
At $\xi=0$, this recovers the original performance-difference estimate, while $\xi=1$ gives the average discounted local improvement along the suffix. Values $0<\xi<1$ balance total improvement against suffix length. Specific domain settings are given in Appendix~\ref{app:experiment-setup}.

\subsection{On-Policy Parallel Tree Search}
\label{app:opts-construction}

A greedy path starts at the root and repeatedly follows $c^\star(p)\in\arg\max_{c\in\mathcal C(p)}\widehat A_c$ until reaching a leaf. The OPTS portion of Figure~\ref{fig:opts-ttpo-overview} illustrates the search process, and Algorithm~\ref{alg:opts-search} gives the batch-level OPTS procedure. It backs up Equation~\ref{eq:appendix-estimated-perf-diff} along each greedy reference path and applies Equation~\ref{eq:appendix-length-penalized-estimate} before selecting the rebranching position.

Selected suffixes in Algorithm~\ref{alg:opts-search} are sampled from $\pi$ at the selected states, while fresh root trajectories that fill unused batch slots are sampled from $\rho_0$; both are generated in one batched rollout. Thus adaptive search requires no separate action-distribution importance correction. The selected state, however, is a function of the observed tree through the backed-up advantages and performance-difference estimates. Consequently, the induced tree distribution generally violates the prefix-measurability condition of Lemma~1. Uniform branch weights still account for branch multiplicity, but they do not by themselves remove this posterior selection effect; Appendix~\ref{app:search-gradient-pg-bias} bounds the resulting gradient bias.

\begin{algorithm}[H]
\caption{On-Policy Parallel Tree Search (OPTS)}
\label{alg:opts-search}
\begin{algorithmic}[1]
\Require Policy $\pi$, environment $P$, initial distribution $\rho_0$, value function $\widehat V$, batch size $B$, batch runs $R$, per-tree search budget $S_{\max}$, budget exponent $\xi$, baseline
\State $s_{i,0}\overset{\mathrm{i.i.d.}}{\sim}\rho_0,\quad i=1,\ldots,B$
\State $\{\tau_i\}\gets\Call{BatchRollout}{\{s_{i,0}\},\pi,P}$
\State $\mathcal T_i\gets\Call{InitTree}{\tau_i},\quad j_i\gets0,\quad i=1,\ldots,B$
\State \Call{TreeGAEBackup}{$\{\tau_i\}\rightarrow\text{roots}$, rewards, $\widehat V$}
\For{$\ell=1,\ldots,R-1$}
    \ForAll{active trees $i$ with $j_i<S_{\max}$ \textbf{in parallel}}
        \State $\tau_i^\star\gets\Call{GreedyPath}{\mathcal T_i}$
        \State $\widehat\Delta(\cdot;\tau_i^\star)\gets\Call{DeltaBackup}{\tau_i^\star,\widehat A,\gamma}$
        \State $s_i^\star\in\arg\max_{s_{i,t}\in\tau_i^\star}\widehat\Delta^{(\xi)}(s_{i,t};\tau_i^\star)$
    \EndFor
    \State $\{s_i^\star\}\gets\operatorname{Take}_B\!\left(\left\{s_i^\star\mid\widehat\Delta^{(\xi)}(s_i^\star;\tau_i^\star)>\mathrm{baseline}\right\}\right)$
    \State $s_{k,0}\overset{\mathrm{i.i.d.}}{\sim}\rho_0,\quad |\{s_{k,0}\}|=B-|\{s_i^\star\}|$
    \State $(\{\tau_i\},\{\tau_k\})\gets\Call{BatchRollout}{\{s_i^\star\}\cup\{s_{k,0}\},\pi,P}$
    \State $\mathcal T_i\gets\Call{Attach}{\mathcal T_i,s_i^\star,\tau_i},\quad j_i\gets j_i+1$
    \State $\mathcal T_k\gets\Call{InitTree}{\tau_k},\quad j_k\gets0$
    \State \Call{TreeGAEBackup}{$\{\tau_i\}\cup\{\tau_k\}\rightarrow\text{roots}$, rewards, $\widehat V$}
\EndFor
\State \Return $\{\mathcal T_i\}$
\end{algorithmic}
\end{algorithm}

Here $k$ indexes new trees, and $\operatorname{Take}_B$ retains at most $B$ states.

\noindent\textbf{Tree-conditioned greedy path.} For a fixed root $s_o$, couple the search trees from one OPTS run as
$\mathcal T_j\sim q_j^{\mathrm{OPTS}}(\cdot\mid s_o;\pi,P)$,
where $j\in\{0,\ldots,S_{\max}\}$ is the number of completed rebranching expansions. Let $\mathcal P_j$ be the retained root-to-leaf paths of $\mathcal T_j$. OPTS only attaches new suffixes and retains all existing paths; after termination, let subsequent trees remain unchanged. Hence, for $0\le j_1\le j_2\le S_{\max}$,
\begin{equation}
\label{eq:appendix-nested-search-trees}
\mathcal T_{j_1}\subseteq\mathcal T_{j_2},
\qquad
\mathcal P_{j_1}
\subseteq
\mathcal P_{j_2}.
\end{equation}
At $j=0$, $\mathcal T_0$ contains the initial path $\tau^{(0)}\sim p_{\pi,P}(\cdot\mid s_o)$ and $\mathcal P_0:=\{\tau^{(0)}\}$. Once the tree is realized, its backed-up scores determine the greedy path
\begin{equation}
\label{eq:appendix-opts-greedy-path}
\tau_j^\star
:=
\Call{GreedyPath}{\mathcal T_j}
\in
\mathcal P_j.
\end{equation}
\noindent\textbf{Greedy path policy.} For a realized tree $\mathcal T_j$, the greedy path policy $\pi_j^\star$ executes the actions recorded on $\tau_j^\star$ while the executed history matches a prefix of that path, and follows $\pi$ after the first deviation.

\noindent\textbf{Induced search policy.} Across OPTS runs, $\mathcal T_j$ is random under $q_j^{\mathrm{OPTS}}$. Let $\tau'$ denote the trajectory generated by executing the greedy path policy from $s_o$. Marginalizing this execution over the latent search-tree randomness defines the induced, generally history-dependent search policy $\pi_j^S$:
\begin{equation}
\label{eq:appendix-induced-search-policy}
p(\tau'\mid\pi_j^S,P,s_o)
:=
\mathbb E_{\mathcal T_j\sim q_j^{\mathrm{OPTS}}}
\left[
p(\tau'\mid\pi_j^\star,P,s_o)
\right].
\end{equation}
Thus $\pi_j^\star$ is conditioned on one realized tree, and $\pi_j^S$ is its marginal execution behavior across OPTS trees.

\subsection{Max-Backup TreeGAE and Search-Budget Improvement}
\label{app:opts-improvement}

In deterministic environments, OPTS uses max-backup TreeGAE throughout search. For $\mathcal C(x)\ne\varnothing$, specialize Equation~\ref{eq:appendix-tree-gae} with
\begin{equation*}
\alpha_{x,x^+}
=
\begin{cases}
1, & x^+=c^\star(x),\\
0, & x^+\ne c^\star(x),
\end{cases}
\qquad
c^\star(x)\in\arg\max_{c\in\mathcal C(x)}
\widehat A_c^{\mathrm{TreeGAE}(V)}.
\end{equation*}
Let $V=V^\pi$ and hold it fixed, giving
\begin{equation}
\widehat A_{\mathrm{max}}(x)
:=
\delta_x^V
+
\gamma\lambda
\max_{c\in\mathcal C(x)}
\widehat A_{\mathrm{max}}(c),
\qquad
\max\varnothing:=0.
\label{eq:appendix-search-max-backup}
\end{equation}
The greedy path follows a maximizing child at each visited state.

For a path $\tau$, define
\begin{equation}
\label{eq:appendix-path-lambda-return}
G_\lambda(\tau)
:=
V^\pi(s_o)
+
\sum_{t=0}^{n-1}
    (\gamma\lambda)^t\delta_t^V.
\end{equation}
For any policy $\mu$, let
\begin{equation}
\label{eq:appendix-lambda-search-objective}
J_\lambda(\mu)
:=
\mathbb E_{\tau\sim\mu}
\left[G_\lambda(\tau)\right].
\end{equation}
At zero search, the only retained path is the original rollout, so $\pi_0^S=\pi$.

\noindent\textbf{Theorem 1 (Monotone Improvement of the Search Objective).} In the deterministic environment above with the fixed exact value function $V=V^\pi$, consider any search with nested path sets (Equation~\ref{eq:appendix-nested-search-trees}) and max-backup greedy selection, and let $\pi_j^S$ be its induced policy as in Equation~\ref{eq:appendix-induced-search-policy}. If $G_\lambda$ is integrable, then for every $\lambda\in[0,1]$ and $0\le j_1\le j_2$,
\begin{equation}
\label{eq:appendix-opts-improvement}
J_\lambda(\pi_{j_2}^S)
\ge
J_\lambda(\pi_{j_1}^S)
\ge
J_\lambda(\pi)
=J(\pi).
\end{equation}
The first inequality is strict if $\Pr\!\left(G_\lambda(\tau_{j_2}^\star)>G_\lambda(\tau_{j_1}^\star)\right)>0$.

\noindent\textbf{Proof.} \textbf{Part I: Path optimality.}

Fix a search tree $\mathcal T_j$ and consider its retained suffix paths. We identify the max-backup values with their maximal discounted TD sums by backward induction.

\emph{Base case.}
At a leaf transition $x$, with $t=d(x)=n-1$,
\begin{equation*}
\max_{\tau:\,\text{suffix from }x}
\sum_{k=t}^{n-1}(\gamma\lambda)^{k-t}\delta_k^V
=\delta_x^V
=\widehat A_{\mathrm{max}}(x).
\end{equation*}

\emph{Induction step.}
Let $t=d(x)<n-1$ and assume the identity holds at every child $c\in\mathcal C(x)$. Partitioning the suffixes by their first child and using $\gamma\lambda\ge0$ gives
\begin{align*}
&\max_{\tau:\,\text{suffix from }x}
\sum_{k=t}^{n-1}(\gamma\lambda)^{k-t}\delta_k^V
\\
={}&\delta_x^V+\gamma\lambda
\max_{c\in\mathcal C(x)}
\left[
\max_{\tau:\,\text{suffix from }c}
\sum_{k=t+1}^{n-1}(\gamma\lambda)^{k-t-1}\delta_k^V
\right]
\\
={}&\delta_x^V+\gamma\lambda
\max_{c\in\mathcal C(x)}\widehat A_{\mathrm{max}}(c)
=\widehat A_{\mathrm{max}}(x).
\end{align*}

\emph{Root conclusion.}
The greedy path attains each recursive maximum. Induction from the leaves to the root therefore yields
\begin{align}
G_\lambda(\tau_j^\star)
&=V^\pi(s_o)+\max_{c\in\mathcal C(o)}\widehat A_{\mathrm{max}}(c)
\nonumber\\
&=V^\pi(s_o)+\max_{\tau\in\mathcal P_j}
\sum_{t=0}^{n-1}(\gamma\lambda)^t\delta_t^V
\nonumber\\
&=\max_{\tau\in\mathcal P_j}G_\lambda(\tau).
\label{eq:appendix-max-backup-lambda-return}
\end{align}
Thus $\tau_j^\star\in\arg\max_{\tau\in\mathcal P_j}G_\lambda(\tau)$.

\textbf{Part II: Budget monotonicity.}\par\nopagebreak[4]

Equations~\ref{eq:appendix-nested-search-trees} and~\ref{eq:appendix-max-backup-lambda-return} give
\begin{align}
G_\lambda(\tau_{j_2}^\star)
&=\max_{\tau\in\mathcal P_{j_2}}G_\lambda(\tau)
\nonumber\\
&\ge\max_{\tau\in\mathcal P_{j_1}}G_\lambda(\tau)
=G_\lambda(\tau_{j_1}^\star)
\nonumber\\
&\ge G_\lambda(\tau^{(0)}).
\label{eq:appendix-lambda-budget-monotonicity-proof}
\end{align}
Taking expectations over the initial state and search tree, Equation~\ref{eq:appendix-induced-search-policy} yields
\begin{align*}
J_\lambda(\pi_{j_2}^S)
&=\mathbb E[G_\lambda(\tau_{j_2}^\star)]
\\
&\ge\mathbb E[G_\lambda(\tau_{j_1}^\star)]
=J_\lambda(\pi_{j_1}^S)
\\
&\ge\mathbb E[G_\lambda(\tau^{(0)})]
=J_\lambda(\pi).
\end{align*}
With $V=V^\pi$, the Bellman equation gives
\begin{align*}
J_\lambda(\pi)
&=\mathbb E_{s_o\sim\rho_0}[V^\pi(s_o)]
+\sum_{t=0}^{n-1}(\gamma\lambda)^t
\underbrace{\mathbb E_{\tau\sim\pi}
\!\left[\mathbb E[\delta_t^V\mid s_t]\right]}_{=0}
\\
&=\mathbb E_{s_o\sim\rho_0}[V^\pi(s_o)]
=J(\pi).
\end{align*}
For strictness,
\begin{equation*}
\Pr\!\left(G_\lambda(\tau_{j_2}^\star)>G_\lambda(\tau_{j_1}^\star)\right)>0
\quad\Longrightarrow\quad
J_\lambda(\pi_{j_2}^S)-J_\lambda(\pi_{j_1}^S)
=\mathbb E[G_\lambda(\tau_{j_2}^\star)-G_\lambda(\tau_{j_1}^\star)]>0.
\end{equation*}
\hfill$\square$

\noindent\textbf{Theorem 2 (Monotone Improvement under OPTS).} In a deterministic environment with the fixed exact value function $V=V^\pi$, assume integrable returns, $\mathrm{baseline}\ge0$, independent complete on-policy suffix rollouts, and suffix-closed admissibility ($s_t$ admissible $\Rightarrow s_{t+1}$ admissible for $t<n-1$). Retain the incumbent on max-backup ties. Then, for $0<\gamma\le1$, $\lambda\in[0,1]$, and $0\le j_1\le j_2\le S_{\max}$,
\begin{equation}
\label{eq:appendix-opts-return-improvement}
J(\pi_{j_2}^S)
\ge
J(\pi_{j_1}^S)
\ge
J(\pi).
\end{equation}

\noindent\textbf{Proof.} \textbf{Part I: Incumbent return bound.}

Fix $\mathcal T_j$, and let $x$ be the transition leaving the selected state $s_t$ on $\tau_j^\star$, with $t=d(x)$. Write $\widehat\Delta(s_k)$ for the estimates on this path in Equation~\ref{eq:appendix-estimated-perf-diff}, with $\widehat\Delta(s_n)=0$. Along the greedy path, Equation~\ref{eq:appendix-search-max-backup} gives
\begin{equation}
G_t-V(s_t)
=
\sum_{k=t}^{n-1}\gamma^{k-t}\delta_k^V
=
-\widehat\Delta(s_t)+\gamma\lambda\widehat\Delta(s_{t+1}).
\label{eq:appendix-opts-incumbent-residual}
\end{equation}
The selection rule gives
\begin{equation*}
\frac{\widehat\Delta(s_t)}{(n-t)^\xi}
>\mathrm{baseline}\ge0
\quad\Longrightarrow\quad
\widehat\Delta(s_t)>0.
\end{equation*}
For $t<n-1$, suffix closure and maximality imply
\begin{equation}
\frac{\widehat\Delta(s_{t+1})}{(n-t-1)^\xi}
\le\frac{\widehat\Delta(s_t)}{(n-t)^\xi}
\quad\Longrightarrow\quad
\widehat\Delta(s_{t+1})
\le
\left(\frac{n-t-1}{n-t}\right)^\xi\widehat\Delta(s_t).
\label{eq:appendix-opts-adjacent-score-order}
\end{equation}
Combining Equations~\ref{eq:appendix-opts-incumbent-residual}--\ref{eq:appendix-opts-adjacent-score-order} yields
\begin{equation}
G_t-V(s_t)
\le
-\left[1-\gamma\lambda
\left(\frac{n-t-1}{n-t}\right)^\xi\right]\widehat\Delta(s_t)
\le0.
\label{eq:appendix-opts-negative-incumbent}
\end{equation}
At the final transition,
\begin{equation*}
t=n-1
\quad\Longrightarrow\quad
G_t-V(s_t)=-\widehat\Delta(s_t)<0.
\end{equation*}

\textbf{Part II: Fresh-suffix return bound.}\par\nopagebreak[4]

For the fresh on-policy suffix $\tau'$, use primes for its sampled quantities and let $\mathcal F_k'$ be the pre-action information. Let $\widehat A_k'$ be its chain GAE from Equation~\ref{eq:appendix-chain-gae}, under the same $V$. On-policy sampling and the Bellman equation give, for every $c\in\mathbb R$,
\begin{equation}
\mathbb E[\widehat A_k'\mid\mathcal F_k']=0
\quad\Longrightarrow\quad
\mathbb E[\widehat A_k'\mathbf 1\{\widehat A_k'>c\}\mid\mathcal F_k']\ge0.
\label{eq:appendix-opts-zero-mean-gae-tail}
\end{equation}
For $0<\lambda\le1$, we prove by backward induction that, for every threshold $c$,
\begin{equation}
\mathbb E\!\left[
\bigl(G_k'-V(s_k')-\widehat A_k'\bigr)
\mathbf 1\{\widehat A_k'>c\}
\mid\mathcal F_k'\right]\ge0.
\label{eq:appendix-opts-return-gae-tail-difference}
\end{equation}

\emph{Base case.}
At the final transition,
\begin{equation*}
G_{n-1}'-V(s_{n-1}')
=\delta_{n-1}'
=\widehat A_{n-1}'.
\end{equation*}

\emph{Induction step.}
Assume Equation~\ref{eq:appendix-opts-return-gae-tail-difference} holds at $k+1$ for every threshold. Subtracting the GAE recursion from the return recursion gives
\begin{align*}
G_k'-V(s_k')-\widehat A_k'
&=\gamma\bigl(G_{k+1}'-V(s_{k+1}')-\lambda\widehat A_{k+1}'\bigr)
\\
&=\gamma\bigl(G_{k+1}'-V(s_{k+1}')-\widehat A_{k+1}'\bigr)
+\gamma(1-\lambda)\widehat A_{k+1}'.
\end{align*}
Given $\mathcal F_{k+1}'$, the first TD residual is known and
\begin{equation*}
\widehat A_k'>c
\quad\Longleftrightarrow\quad
\widehat A_{k+1}'>\frac{c-\delta_k'}{\gamma\lambda}.
\end{equation*}
The induction hypothesis and Equation~\ref{eq:appendix-opts-zero-mean-gae-tail} therefore yield
\begin{align*}
&\mathbb E\!\left[
\bigl(G_k'-V(s_k')-\widehat A_k'\bigr)
\mathbf 1\{\widehat A_k'>c\}
\mid\mathcal F_{k+1}'\right]
\\
={}&\gamma
\underbrace{\mathbb E\!\left[
\bigl(G_{k+1}'-V(s_{k+1}')-\widehat A_{k+1}'\bigr)
\mathbf 1\{\widehat A_k'>c\}
\mid\mathcal F_{k+1}'\right]}_{\ge0}
\\
&{}+\gamma(1-\lambda)
\underbrace{\mathbb E\!\left[
\widehat A_{k+1}'\mathbf 1\{\widehat A_k'>c\}
\mid\mathcal F_{k+1}'\right]}_{\ge0}
\ge0.
\end{align*}
Taking $\mathbb E[\,\cdot\mid\mathcal F_k']$ completes the induction.

\emph{Initial-state conclusion.}
Equations~\ref{eq:appendix-opts-zero-mean-gae-tail}--\ref{eq:appendix-opts-return-gae-tail-difference} give
\begin{equation}
\mathbb E[(G_t'-V(s_t))\mathbf 1\{\widehat A_t'>c\}\mid s_t]
\ge
\mathbb E[\widehat A_t'\mathbf 1\{\widehat A_t'>c\}\mid s_t]
\ge0.
\label{eq:appendix-opts-upper-tail-return}
\end{equation}
At $\lambda=0$, $\widehat A_t'=\delta_t'$ and the later TD residuals have zero conditional contribution, so
\begin{equation*}
\mathbb E[(G_t'-V(s_t))\mathbf 1\{\widehat A_t'>c\}\mid s_t]
=\mathbb E[\delta_t'\mathbf 1\{\delta_t'>c\}\mid s_t]
\ge0.
\end{equation*}

\textbf{Part III: Budget monotonicity.}

The new suffix is selected when $\widehat A_t'>\widehat A_{\max}(x)$, with the greedy prefix unchanged. Conditional on $\mathcal T_j$, independence of the fresh rollout and Equations~\ref{eq:appendix-opts-negative-incumbent} and~\ref{eq:appendix-opts-upper-tail-return} give
\begin{align}
&\gamma^t
\mathbb E\!\left[
(G_t'-G_t)\mathbf 1\{\widehat A_t'>\widehat A_{\max}(x)\}
\mid\mathcal T_j
\right]
\nonumber\\
={}&\gamma^t\mathbb E\!\left[
(G_t'-V(s_t))\mathbf 1\{\widehat A_t'>\widehat A_{\max}(x)\}
\mid\mathcal T_j\right]
\nonumber\\
&{}+\gamma^t\bigl(V(s_t)-G_t\bigr)
\Pr\!\left(\widehat A_t'>\widehat A_{\max}(x)\mid\mathcal T_j\right)
\nonumber\\
\ge{}&
\gamma^t\bigl(V(s_t)-G_t\bigr)
\Pr\!\left(\widehat A_t'>\widehat A_{\max}(x)\mid\mathcal T_j\right)
\ge0.
\label{eq:appendix-opts-one-step-return-improvement}
\end{align}
Terminated runs contribute zero increment. The tower property yields
\begin{align*}
&J(\pi_{j+1}^S)-J(\pi_j^S)
\\
={}&\mathbb E_{s_o\sim\rho_0,\mathcal T_j}\!\left[
\gamma^t\mathbb E\!\left[
(G_t'-G_t)\mathbf 1\{\widehat A_t'>\widehat A_{\max}(x)\}
\mid\mathcal T_j\right]\right]
\ge0.
\end{align*}
Iterating from $\pi_0^S=\pi$ proves Equation~\ref{eq:appendix-opts-return-improvement}.
For strictness at an expansion $j$,
\begin{equation*}
\Pr\!\left(G_t<V(s_t),\ \widehat A_t'>\widehat A_{\max}(x)\right)>0
\quad\Longrightarrow\quad
J(\pi_{j+1}^S)>J(\pi_j^S).
\end{equation*}
\hfill$\square$

At $\lambda=0$, OPTS selects the greedy-$Q^\pi$ path; at $\lambda=1$, it selects the largest-return path. Intermediate values trade off preserving the best complete return already found against selecting high-$Q^\pi$ states from which subsequent rebranching can improve it.

\section{OPTS-TTPO: Search-Enhanced Policy Gradients}
\label{app:opts-ttpo-theory}

The Branch Aggregation Lemma is the foundation for learning from prefix-measurable trees, and the performance-difference estimate allocates rollout budget. OPTS deliberately departs from the lemma's sampling condition by using observed suffixes to decide where to expand and, under max backup, which discovered suffix to propagate. Each new suffix remains conditionally sampled from the current policy, but the posterior decisions change the induced tree distribution. This section explains the structure contributed by that posterior information and bounds the resulting gradient bias.

We use uniform local branch weights $\alpha_{p,c}=1/|\mathcal C(p)|$, with $W$ given by Equation~\ref{eq:appendix-global-branch-weight}, to account for branch multiplicity. For a fixed state $s_t$ and action sequence $a_{t:n-1}$, deterministic dynamics uniquely determine the suffix return $G_t$; replaying the same sequence therefore reproduces its return. In this setting, we use max-backup TreeGAE, the backup rule under which Appendix~\ref{app:opts-theory} establishes the search-improvement guarantees. Under stochastic dynamics, $G_t$ need not be uniquely determined even after conditioning on the same action sequence. Maximizing a finite set of backed-up suffix scores can then select both suffix quality and environmental noise, causing high-variance return distributions to be favored over suffixes with higher conditional expected returns. We therefore use mean-backup TreeGAE to average sampled suffixes and reduce this max-selection effect while retaining the coverage gained from branching. Adaptive expansion affects both variants; max backup contributes the additional prefix-credit term analyzed below.

\subsection{Posterior Max Backup as Prefix Credit}
\label{app:posterior-prefix-credit}

Fix a tree $\mathcal T$ sampled from $\pi_{\theta_{\mathrm{old}}}$, a critic $V$, and a common $\lambda$. Using this critic, compute $\widehat A_{\max}$ by Equation~\ref{eq:appendix-search-max-backup} and $\widehat A_{\mathrm{mean}}$ by Equation~\ref{eq:appendix-tree-gae} with the uniform weights above. Let $\widehat g_{\max}$ and $\widehat g_{\mathrm{mean}}$ denote their TTPG directions, evaluated at $\theta_{\mathrm{old}}$:
\begin{equation}
\widehat g_{\max}
:=
\left.
\sum_{x\in\mathcal T}
W(x)\gamma^{d(x)}
\widehat A_{\max}(x)
\nabla_\theta\log\pi_\theta(a_x\mid s_x)
\right|_{\theta=\theta_{\mathrm{old}}},
\label{eq:appendix-search-gradient-definition}
\end{equation}
with $\widehat g_{\mathrm{mean}}$ defined analogously. The tree, critic, weights, and backed-up marks are held fixed during this actor update.

At a transition $x$ with children, define the best-minus-average suffix gap
\begin{equation}
b(x)
:=
\max_{c\in\mathcal C(x)}\widehat A_{\max}(c)
-
\sum_{c\in\mathcal C(x)}\alpha_{x,c}\widehat A_{\max}(c)
\ge0,
\label{eq:appendix-posterior-suffix-gap}
\end{equation}
and set $b(x)=0$ at a leaf. Write $x\preceq u$ when $x$ lies on the root-to-$u$ path, including $u$ itself.

\noindent\textbf{Theorem 3 (Posterior Prefix Reinforcement).} For every such tree, define the frozen-tree surrogate
\begin{equation}
\mathcal L_{\mathrm{search}}(\theta)
:=
\gamma\lambda
\sum_{u\in\mathcal T}
W(u)\gamma^{d(u)}b(u)
\sum_{x\preceq u}
\lambda^{d(u)-d(x)}
\log\pi_\theta(a_x\mid s_x).
\label{eq:appendix-prefix-reinforcement-objective}
\end{equation}
Then
\begin{equation}
\widehat g_{\max}
=
\widehat g_{\mathrm{mean}}
+
\left.
\nabla_\theta\mathcal L_{\mathrm{search}}(\theta)
\right|_{\theta=\theta_{\mathrm{old}}}.
\label{eq:appendix-max-mean-gradient-decomposition}
\end{equation}
Consequently, at $\theta_{\mathrm{old}}$, their directional derivatives of this surrogate satisfy
\begin{equation}
\nabla_\theta\mathcal L_{\mathrm{search}}^\top\widehat g_{\max}
-
\nabla_\theta\mathcal L_{\mathrm{search}}^\top\widehat g_{\mathrm{mean}}
=
\|\nabla_\theta\mathcal L_{\mathrm{search}}\|^2
\ge0.
\label{eq:appendix-relative-prefix-reinforcement}
\end{equation}

\noindent\textbf{Proof.} \textbf{Part I: Propagation of the suffix gap.}
Subtracting the two backup recursions gives
\begin{align}
&
\widehat A_{\max}(x)-\widehat A_{\mathrm{mean}}(x)
\nonumber\\
={}&
\gamma\lambda\left[
b(x)+
\sum_{c\in\mathcal C(x)}\alpha_{x,c}
\bigl(\widehat A_{\max}(c)-\widehat A_{\mathrm{mean}}(c)\bigr)
\right].
\label{eq:appendix-backup-gap-recursion}
\end{align}
Expanding to the leaves and using the product definition of $W$ yields
\begin{equation}
W(x)\bigl(\widehat A_{\max}(x)-\widehat A_{\mathrm{mean}}(x)\bigr)
=
\gamma\lambda
\sum_{u:\,x\preceq u}
W(u)(\gamma\lambda)^{d(u)-d(x)}b(u).
\label{eq:appendix-backup-gap-expansion}
\end{equation}

\textbf{Part II: Credit assigned to the prefix.}
Substitute Equation~\ref{eq:appendix-backup-gap-expansion} into the difference of the TTPG sums and exchange the ancestor--descendant sums:
\begin{align}
&
\widehat g_{\max}-\widehat g_{\mathrm{mean}}
\nonumber\\
={}&
\left.
\gamma\lambda
\sum_{u\in\mathcal T}
W(u)\gamma^{d(u)}b(u)
\sum_{x\preceq u}
\lambda^{d(u)-d(x)}
\nabla_\theta\log\pi_\theta(a_x\mid s_x)
\right|_{\theta=\theta_{\mathrm{old}}}.
\label{eq:appendix-prefix-credit-expansion}
\end{align}
This is the gradient of Equation~\ref{eq:appendix-prefix-reinforcement-objective}. Taking its inner product with that gradient proves Equation~\ref{eq:appendix-relative-prefix-reinforcement}.
\hfill$\square$

\noindent\textbf{What posterior max backup contributes.} In a deterministic environment with $V=V^\pi$, Equation~\ref{eq:appendix-max-backup-lambda-return} identifies each max-backup mark with the best retained suffix $\lambda$-return relative to its source-state value. Thus $b(x)$ measures the best-minus-average $\lambda$-return gap among the retained child suffixes. Max backup passes this relative gain to the actions leading to the branch point. At $\lambda=1$, the inner sum in Equation~\ref{eq:appendix-prefix-reinforcement-objective} is the log-likelihood of the action prefix through $u$; for $0<\lambda<1$, the additional credit decays with the distance to that branch point. At $\lambda=0$, both backups reduce to the local TD residual.

Equation~\ref{eq:appendix-relative-prefix-reinforcement} quantifies the additional first-order change in this weighted prefix objective under a max-backup update. It explains how a discovered suffix can influence upstream learning before the current policy reliably reproduces that suffix. Subsequent on-policy rollouts and downstream updates determine how this credit is translated into realized return.

The identity is pathwise, so it also holds after averaging the two directions over the same OPTS tree distribution. Adaptive expansion determines which suffixes are available; posterior max backup determines how their relative values are propagated. We account for both effects below.

\subsection{Controlling the Search-Induced Gradient Bias}
\label{app:search-gradient-pg-bias}

Draw an initial trajectory $\tau^{(0)}\sim\pi_{\theta_{\mathrm{old}}}$ and let $j\in\{0,\ldots,S_{\max}\}$ be its random number of OPTS expansions. Allocation and stopping may depend on all observed search outcomes. Evaluate $\widehat g_{\max}$ and $\widehat g_{\mathrm{mean}}$ on the resulting tree $\mathcal T_j$, and write $\overline g_{\max}:=\mathbb E[\widehat g_{\max}]$. The mean-backup direction contains the adaptive tree-selection effect; the difference $\widehat g_{\max}-\widehat g_{\mathrm{mean}}$ isolates the additional prefix-credit contribution of posterior max backup.
Let $\widehat g_{\mathrm{ch}}(\tau)$ be the chain-GAE policy-gradient sum with the same critic and discount, and define
$\varepsilon_0:=\|\mathbb E[\widehat g_{\mathrm{ch}}(\tau^{(0)})]-\nabla J(\theta_{\mathrm{old}})\|$.
This initial-chain bias vanishes with exact advantages. For chain GAE, exact $V^{\pi_{\theta_{\mathrm{old}}}}$ likewise gives
\[
\mathbb E[\widehat g_{\mathrm{ch}}(\tau^{(0)})]
=\nabla J(\theta_{\mathrm{old}}),
\qquad \varepsilon_0=0,
\]
by the chain policy-gradient and GAE identities.

Suppose $|r|\le R_{\max}$, $|V|\le V_{\max}$, and
$\|\left.\nabla_\theta\log\pi_\theta(a\mid s)\right|_{\theta=\theta_{\mathrm{old}}}\|\le L_\pi$.
The TD residual satisfies
\[
|r+\gamma V(s')-V(s)|
\le |r|+\gamma|V(s')|+|V(s)|
\le R_{\max}+(1+\gamma)V_{\max}.
\]
For either tree backup,
\[
\left|\sum_c\alpha_{x,c}\widehat A(c)\right|
\le\sum_c\alpha_{x,c}|\widehat A(c)|
\le\max_c|\widehat A(c)|,
\qquad
\left|\max_c\widehat A(c)\right|
\le\max_c|\widehat A(c)|.
\]
Recursing from the leaves over at most $n$ steps gives the uniform absolute bound for chain GAE and both tree backups:
\begin{equation}
A_{\mathrm{bd}}
:=
\bigl[R_{\max}+(1+\gamma)V_{\max}\bigr]
\sum_{t=0}^{n-1}(\gamma\lambda)^t.
\label{eq:appendix-treegae-mark-bound}
\end{equation}
With $C_n:=\sum_{t=0}^{n-1}\gamma^t$, the normalized branch weights have total mass one at each depth, so every corresponding tree or chain direction has norm at most $C_nA_{\mathrm{bd}}L_\pi$.

Let
\begin{equation}
p:=\Pr(j>0),
\qquad
\rho_S:=\mathbb E[1-2^{-j}]
\le p(1-2^{-S_{\max}}).
\label{eq:appendix-search-population-gradient}
\end{equation}
Here $p$ is the population fraction of root trees receiving at least one search expansion, and $\rho_S$ bounds the expected total path weight assigned to paths other than the initial trajectory. The extra population-gradient contribution of max backup is measured by
\begin{equation}
\varepsilon_{\mathrm{backup}}
:=
\left\|
\mathbb E\left[
\widehat g_{\max}
-\widehat g_{\mathrm{mean}}
\right]
\right\|.
\label{eq:appendix-max-backup-bias}
\end{equation}

\noindent\textbf{Proposition 2 (Search-Induced Gradient Bias Bound).} Under the preceding conditions,
\begin{align}
&
\|\overline g_{\max}-\nabla J(\theta_{\mathrm{old}})\|
\nonumber\\
\le{}&
\varepsilon_0+
\min\left\{
2pC_nA_{\mathrm{bd}}L_\pi,\;
2\rho_SC_nA_{\mathrm{bd}}L_\pi+\varepsilon_{\mathrm{backup}}
\right\},
\label{eq:appendix-search-gradient-bias-bound}
\end{align}
where the backup contribution satisfies
\begin{equation}
\varepsilon_{\mathrm{backup}}
\le
\gamma\lambda L_\pi
\mathbb E\left[
\sum_{u\in\mathcal T_j}
W(u)\gamma^{d(u)}b(u)
\sum_{r=0}^{d(u)}\lambda^r
\right].
\label{eq:appendix-gap-dependent-bias-bound}
\end{equation}

\noindent\textbf{Proof.} \textbf{Part I: Adaptive sampling and branch weighting.}
For each retained path $\tau=(x_0,\ldots,x_{n-1})\in\mathcal P_j$, let $w_\tau:=W(x_{n-1})$. Expanding the mean recursion gives the pathwise identity
\begin{equation}
\widehat g_{\mathrm{mean}}
=
\sum_{\tau\in\mathcal P_j}
w_\tau\widehat g_{\mathrm{ch}}(\tau),
\qquad
\sum_\tau w_\tau=1.
\label{eq:appendix-mean-gradient-leaf-mixture}
\end{equation}
This uses the direct--recursive aggregation identity from Lemma~1. Adding one child to a branch point with $m$ children on the initial path multiplies its existing local weight by $m/(m+1)\ge1/2$. Expansions elsewhere leave that path's weight unchanged. Hence $w_{\tau^{(0)}}\ge2^{-j}$ and
\begin{align}
&
\left\|
\widehat g_{\mathrm{mean}}
-\widehat g_{\mathrm{ch}}(\tau^{(0)})
\right\|
\nonumber\\
= {}&
\left\|
\sum_\tau w_\tau
\left[\widehat g_{\mathrm{ch}}(\tau)
-\widehat g_{\mathrm{ch}}(\tau^{(0)})\right]
\right\|
\nonumber\\
= {}&
\left\|
\sum_{\tau\ne\tau^{(0)}}w_\tau
\left[\widehat g_{\mathrm{ch}}(\tau)
-\widehat g_{\mathrm{ch}}(\tau^{(0)})\right]
\right\|
\nonumber\\
\le{}&
\sum_{\tau\ne\tau^{(0)}}w_\tau
\left\|\widehat g_{\mathrm{ch}}(\tau)
-\widehat g_{\mathrm{ch}}(\tau^{(0)})\right\|
\nonumber\\
\le{}&
2C_nA_{\mathrm{bd}}L_\pi(1-w_{\tau^{(0)}}).
\label{eq:appendix-tree-selection-bias-bound}
\end{align}
The initial trajectory retains its on-policy marginal law. By the triangle inequality and $\|\mathbb E[\cdot]\|\le\mathbb E[\|\cdot\|]$,
\begin{align}
&
\left\|
\mathbb E[\widehat g_{\mathrm{mean}}]
-\nabla J(\theta_{\mathrm{old}})
\right\|
\nonumber\\
={}&
\Bigl\|
\mathbb E[\widehat g_{\mathrm{mean}}-\widehat g_{\mathrm{ch}}(\tau^{(0)})]
+\mathbb E[\widehat g_{\mathrm{ch}}(\tau^{(0)})]
-\nabla J(\theta_{\mathrm{old}})
\Bigr\|
\nonumber\\
\le{}&
\varepsilon_0+
\left\|\mathbb E[\widehat g_{\mathrm{mean}}-\widehat g_{\mathrm{ch}}(\tau^{(0)})]\right\|
\nonumber\\
\le{}&
\varepsilon_0+
\mathbb E\!\left[\left\|\widehat g_{\mathrm{mean}}-\widehat g_{\mathrm{ch}}(\tau^{(0)})\right\|\right]
\nonumber\\
\le{}&
\varepsilon_0+2C_nA_{\mathrm{bd}}L_\pi\,\mathbb E[1-w_{\tau^{(0)}}]
\nonumber\\
\le{}&
\varepsilon_0+2C_nA_{\mathrm{bd}}L_\pi\,\mathbb E[1-2^{-j}]
\nonumber\\
={}&
\varepsilon_0+2\rho_SC_nA_{\mathrm{bd}}L_\pi.
\label{eq:appendix-search-selection-bias-bound}
\end{align}

\textbf{Part II: Posterior max backup.}
Equation~\ref{eq:appendix-max-mean-gradient-decomposition} separates the two contributions exactly:
\begin{align}
&
\overline g_{\max}-\nabla J(\theta_{\mathrm{old}})
\nonumber\\
={}&
\mathbb E[\widehat g_{\mathrm{mean}}]
-\nabla J(\theta_{\mathrm{old}})
\nonumber\\
&{}+
\mathbb E[\widehat g_{\max}
-\widehat g_{\mathrm{mean}}].
\label{eq:appendix-total-search-bias-decomposition}
\end{align}
By the triangle inequality and Part I,
\begin{align*}
&\|\overline g_{\max}-\nabla J(\theta_{\mathrm{old}})\|\\
\le{}&
\left\|\mathbb E[\widehat g_{\mathrm{mean}}]-\nabla J(\theta_{\mathrm{old}})\right\|
+\left\|\mathbb E[\widehat g_{\max}-\widehat g_{\mathrm{mean}}]\right\|\\
\le{}&
\varepsilon_0+2\rho_SC_nA_{\mathrm{bd}}L_\pi
+\varepsilon_{\mathrm{backup}},
\end{align*}
which gives the second alternative in Equation~\ref{eq:appendix-search-gradient-bias-bound}.
Using Equation~\ref{eq:appendix-prefix-credit-expansion} and the nonnegative coefficients,
\begin{align*}
\varepsilon_{\mathrm{backup}}
={}&\left\|\mathbb E[\widehat g_{\max}-\widehat g_{\mathrm{mean}}]\right\|\\
\le{}&\mathbb E\!\left[\|\widehat g_{\max}-\widehat g_{\mathrm{mean}}\|\right]\\
={}&\gamma\lambda\,
\mathbb E\!\left[
\left\|
\sum_{u\in\mathcal T_j}W(u)\gamma^{d(u)}b(u)
\sum_{x\preceq u}\lambda^{d(u)-d(x)}
\left.\nabla_\theta\log\pi_\theta(a_x\mid s_x)\right|_{\theta=\theta_{\mathrm{old}}}
\right\|
\right]\\
\le{}&\gamma\lambda\,
\mathbb E\!\left[
\sum_{u\in\mathcal T_j}W(u)\gamma^{d(u)}b(u)
\sum_{x\preceq u}\lambda^{d(u)-d(x)}
\left\|\left.\nabla_\theta\log\pi_\theta(a_x\mid s_x)\right|_{\theta=\theta_{\mathrm{old}}}\right\|
\right]\\
\le{}&\gamma\lambda L_\pi\,
\mathbb E\!\left[
\sum_{u\in\mathcal T_j}W(u)\gamma^{d(u)}b(u)
\sum_{x\preceq u}\lambda^{d(u)-d(x)}
\right]\\
={}&\gamma\lambda L_\pi\,
\mathbb E\!\left[
\sum_{u\in\mathcal T_j}W(u)\gamma^{d(u)}b(u)
\sum_{r=0}^{d(u)}\lambda^r
\right],
\end{align*}
which proves Equation~\ref{eq:appendix-gap-dependent-bias-bound}.

\textbf{Part III: A uniform cap.}
On $j=0$, the max-backup direction equals the initial chain direction. Consequently,
\begin{align}
&
\|\overline g_{\max}-\nabla J(\theta_{\mathrm{old}})\|
\nonumber\\
\le{}&
\varepsilon_0+
\mathbb E\left[
\mathbf 1\{j>0\}
\left\|\widehat g_{\max}
-\widehat g_{\mathrm{ch}}(\tau^{(0)})\right\|
\right]
\nonumber\\
\le{}&
\varepsilon_0+2pC_nA_{\mathrm{bd}}L_\pi,
\end{align}
which proves the first alternative.
\hfill$\square$

\noindent\textbf{Dependence on the search budget.} Both search contributions vanish on unexpanded trees. Since $|\widehat A_{\max}(c)|\le A_{\mathrm{bd}}$,
\[
0\le b(u)
\le A_{\mathrm{bd}}-(-A_{\mathrm{bd}})
=2A_{\mathrm{bd}}.
\]
A node with one child has $b(u)=0$, and each expansion adds one suffix. Thus
\[
\bigl|\{u\in\mathcal T_j:b(u)>0\}\bigr|\le j,
\qquad
\sum_{u:\,b(u)>0}W(u)\gamma^{d(u)}\le j.
\]
The unit branch mass at each depth also gives
\[
\sum_{u:\,b(u)>0}W(u)\gamma^{d(u)}
\le\sum_{t=0}^{n-1}\gamma^t
\sum_{u:\,d(u)=t}W(u)
=C_n.
\]
Combining these bounds with Equation~\ref{eq:appendix-gap-dependent-bias-bound},
\begin{align}
\varepsilon_{\mathrm{backup}}
\le{}&
2\gamma\lambda A_{\mathrm{bd}}L_\pi\,
\mathbb E\!\left[\sum_{u:\,b(u)>0}W(u)\gamma^{d(u)}\right]
\sum_{r=0}^{n-1}\lambda^r
\nonumber\\
\le{}&
2\gamma\lambda A_{\mathrm{bd}}L_\pi\,
\mathbb E[\min\{j,C_n\}]
\sum_{r=0}^{n-1}\lambda^r
\nonumber\\
\le{}&
2\gamma\lambda A_{\mathrm{bd}}L_\pi\,
\mathbb E[\mathbf 1\{j>0\}]\min\{S_{\max},C_n\}
\sum_{r=0}^{n-1}\lambda^r
\nonumber\\
={}&
2p\gamma\lambda A_{\mathrm{bd}}L_\pi
\min\{S_{\max},C_n\}
\sum_{r=0}^{n-1}\lambda^r.
\label{eq:appendix-budget-backup-bias-bound}
\end{align}
Thus the searched-tree fraction scales both contributions, the expansion cap bounds the number of posterior decisions, and the suffix gaps quantify the additional credit actually introduced by max backup. Normalized branch weights keep the total gradient scale bounded as the tree grows. At $S_{\max}=0$, the bound reduces to $\varepsilon_0$, and exact values recover the unbiased chain policy gradient.

\subsection{OPTS-TTPO Training Procedure}

\begin{algorithm}[H]
\caption{OPTS-TTPO training step}
\label{alg:opts-ttpo}
\begin{algorithmic}[1]
\Require $\pi_\theta$, $P$, $\rho_0$, $\widehat V$, $B$, $R$, $S_{\max}$, $\xi$, baseline
\State $\mathcal B\gets\Call{OPTS}{\pi_\theta,P,\rho_0,\widehat V,B,R,S_{\max},\xi,\mathrm{baseline}}$ by Algorithm~\ref{alg:opts-search}
\State Set $\alpha_{p,c}\gets1/|\mathcal C(p)|$ and compute $W(x)$ by Equation~\ref{eq:appendix-global-branch-weight}
\State Construct TTPO losses using the TreeGAE advantages in $\mathcal B$ and branch weights $W(x)$ by Section~\ref{sec:ttpo-objective}
\State Update the critic with $\mathcal{L}_V^{\mathrm{TTPO}}$ and the actor with $\mathcal{L}_\pi^{\mathrm{TTPO}}$
\end{algorithmic}
\end{algorithm}

%% file: sections/Appendix_experiments.tex
\section{Experimental Setup}
\label{app:experiment-setup}

We describe the protocols in the same order as the main text. Shared optimization details, data-overlap auditing, prompt formatting, and the language-model MDP formulation follow these experiment-specific settings.

\subsection{Tree-Gradient Estimation: Setup and Protocol}
\label{app:rq1-gradient-protocol}

For the gradient-aggregation experiment in Section~\ref{sec:tree-policy-gradient-evaluation}, we freeze a step-400 \texttt{Qwen3-1.7B} PPO actor and use 16,384 training prompts. The estimator side contains eight midpoint-branch tree groups with $K\in\{0,1,3,7,15\}$ additional suffixes; the reference side contains 32 independently sampled chain groups. Both sides use global token-level aggregation. The reference gradient, tree construction, and metrics are described below.

\noindent\textbf{Reference gradient.} We freeze the step-400 \texttt{Qwen3-1.7B} PPO checkpoint and evaluate all gradients on the 16,384-prompt training set. Returns are binary verifier outcomes, with $\gamma=\lambda=1$ and no learned baseline. For a collection $\mathcal C$ of chain responses, define
\begin{equation}
N(\mathcal C)
:=
\sum_{\tau\in\mathcal C}\sum_{x\in\tau}
R(\tau)\nabla_\theta\log\pi_\theta(a_x\mid s_x),
\qquad
D(\mathcal C)
:=
\sum_{\tau\in\mathcal C}|\tau|,
\end{equation}
where $x$ ranges over valid response-token transitions. We independently sample 32 chain groups, each containing one response per prompt, and form the reference over their union $\mathcal C^\star$ as $g^\star=N(\mathcal C^\star)/D(\mathcal C^\star)$. This reference is finite; reported bias is deviation from $g^\star$, not population bias.

\noindent\textbf{Tree gradients.} We independently sample eight tree groups, each containing one tree per prompt. Each tree branches at the midpoint of its backbone response, retains the original suffix, and samples 15 additional on-policy suffixes. The settings $K\in\{0,1,3,7,15\}$ use nested subsets of the first $K$ additional suffixes. Within a tree, let $R_j$ be the binary return of suffix $j\in\{0,\ldots,K\}$. The shared-prefix target is the branch mean $\bar R_K:=(K+1)^{-1}\sum_{j=0}^{K}R_j$, while every token on suffix $j$ uses $R_j$.

For $A\in\{\mathrm{NaivePG},\mathrm{TTPG}\}$, let $w^A(x)=1$ for NaivePG and $w^A(x)=W(x)$ for TTPG. Under the single uniform branch point used here, $W(x)=1$ on the shared prefix and $W(x)=1/(K+1)$ on every suffix. Writing $\widetilde R_K(x)$ for the assigned prefix or suffix target of $x$ and $\mathcal X_{i,K}$ for all valid response-token occurrences in group $i$, we accumulate
\begin{equation}
N^A_{i,K}
:=
\sum_{x\in\mathcal X_{i,K}}
w^A(x)\widetilde R_K(x)
\nabla_\theta\log\pi_\theta(a_x\mid s_x),
\qquad
D^A_{i,K}
:=
\sum_{x\in\mathcal X_{i,K}}w^A(x).
\end{equation}
Every $M\in\{1,2,4\}$ consecutive groups form one non-overlapping block $\mathcal B_r$ and one globally normalized estimate
\begin{equation}
\hat g^A_{r,K,M}
=
\frac{\sum_{i\in\mathcal B_r}N^A_{i,K}}
{\sum_{i\in\mathcal B_r}D^A_{i,K}},
\qquad
|\mathcal B_r|=M,
\qquad
R_M=\frac{8}{M}.
\label{eq:rq1-global-token-estimator}
\end{equation}
Thus each estimate merges group numerators and token-weight masses before division.

\noindent\textbf{Metrics.} For each $A$, $K$, and $M$, write $g_r=\hat g^A_{r,K,M}$ and $\bar g=R_M^{-1}\sum_{r=1}^{R_M}g_r$. We report
\begin{align}
\mathrm{Bias}&=\frac{\lVert\bar g-g^\star\rVert_2}{\lVert g^\star\rVert_2}, &
\mathrm{Var}&=\frac{\sum_{r=1}^{R_M}\lVert g_r-\bar g\rVert_2^2}{(R_M-1)\lVert g^\star\rVert_2^2},
\label{eq:rq1-bias-variance}
\\
\mathrm{MSE}&=\frac{\sum_{r=1}^{R_M}\lVert g_r-g^\star\rVert_2^2}{R_M\lVert g^\star\rVert_2^2}, &
\mathrm{Cos}&=\frac{1}{R_M}\sum_{r=1}^{R_M}\frac{\langle g_r,g^\star\rangle}{\lVert g_r\rVert_2\lVert g^\star\rVert_2}.
\label{eq:rq1-mse-cosine}
\end{align}

\subsection{OPTS Search and Test-Time Scaling}
\label{app:setup-rq2}

\noindent\textbf{Exact-value experiment.} We construct 32 fixed environment--policy configurations (seeds 0--31). Each environment is a depth-4 deterministic binary tree with 15 nonterminal states and 16 terminal states. Intermediate rewards are zero; rewards on terminal transitions are a random permutation of $\{0/15,1/15,\ldots,15/15\}$. At each nonterminal state, the right-action probability is drawn independently and uniformly from $\{1/4,1/3,1/2,2/3,3/4\}$, with the complementary probability assigned to the left action. Reward permutations and policy probabilities use separate random streams. The policy remains fixed, and its exact $V^\pi$ is computed recursively with $\gamma=1$.

Both guidance modes use max-backup TreeGAE, $\xi=0$, a zero search baseline, and $\lambda\in\{0,0.3,0.6,0.95\}$. Reward-guided OPTS uses full advantages including the terminal reward. Value-guided OPTS truncates advantage computation before the terminal transition and bootstraps with exact $V^\pi$ at the last nonterminal state, equivalently setting the final TD residual to zero. Both modes sample complete on-policy suffixes; only the advantage computation is truncated. Each search round selects the largest performance-difference estimate on the current greedy path and samples one new suffix. Search stops if no estimate exceeds zero; max-backup ties retain the incumbent path, and ties between candidate positions select the earliest position.

We enumerate all possible initial trajectories and subsequent suffix samples, weighting them by their policy probabilities to obtain exact expectations for every budget $j\in\{0,1,\ldots,15\}$. Both modes are evaluated using the full-reward $J_\lambda$ and true return $J$ along their guidance-greedy paths. For each mode and metric, the monotonicity check comprises $32\times4\times15=1{,}920$ comparisons between adjacent budgets. Figure~\ref{fig:opts-search-budget-evidence} (top) plots improvements relative to $j=0$; curves and shaded bands show the mean and $\pm$ one sample standard deviation across the 32 configurations.

\noindent\textbf{Learned-critic experiment.} For the search-budget experiment in Section~\ref{sec:opts-search-scaling}, we freeze a step-400 \texttt{Qwen3-1.7B} OPTS-TTPO actor and critic, use 902 held-out prompts with 32 trees per prompt, and vary $S_{\max}\in\{0,1,3,7,15\}$ with $\lambda=0.999$. Reward-guided and value-guided search reuse the same root-response texts. We report the full verifier-reward $J_\lambda$ and terminal verifier return $J$ along the guidance-greedy paths, relative to $S_{\max}=0$.

\noindent\textbf{Matched-budget test-time scaling.} For matched rollout-budget test-time scaling, we evaluate fixed $S_{\max}=3$ at budgets $k\in\{8,16,32,64,128\}$, corresponding to $k\times902$ rollouts. A fresh root rollout and a suffix rollout from a branch point each count as one rollout. Every search round contains 902 rollouts, with unused rebranching slots filled by fresh initial trees. Reward-guided OPTS is compared with i.i.d. \texttt{pass@k}; value-guided OPTS uses the learned value function and truncated GAE and returns the majority answer among value-greedy tree responses, with i.i.d. self-consistency (\texttt{cons@k}) as its baseline. OPTS and i.i.d. baselines use temperature 1.0, top-$p=0.95$, and no top-$k$ truncation.

\subsection{Coverage--Bias and Prefix-Credit Mechanism Protocol}
\label{app:mechanistic-exp}

This protocol produces the two panels in Section~\ref{sec:mechanistic-exp}. We freeze the step-400 \texttt{Qwen3-1.7B} OPTS-TTPO actor--critic checkpoint; it is distinct from the PPO checkpoint used for the branch-aggregation experiment. We run the experiment on 16,384 training prompts. The fixed-branch method branches at response token 128 and skips responses shorter than 128 tokens; the other method runs OPTS.

\noindent\textbf{Matched coverage and bias comparison.} For the OPTS side, we retain eight trees per prompt and record the search snapshots $s\in\{0,1,3,7\}$, with $p=1$. OPTS selects the rebranching trees using its performance-difference rule. The fixed controls branch at response token 128. At every prompt and search round, each fixed control receives exactly the number of branches selected by OPTS for that prompt and round; its fixed-position trees are sampled uniformly without replacement and do not use their own rewards for allocation. The plotted methods are Fixed + TTPG, Fixed + NaivePG, and OPTS + TTPG (the OPTS max-backup snapshot). We use $M=1$ theory aggregation and recompute the gradient coefficients from direct returns with $V=0$, so panel~(a) isolates the coverage--bias relation rather than learned-critic error. The horizontal coordinate is the relative bias difference from $s=0$; the vertical coordinate is the prompt-coverage gain in percentage points.

\noindent\textbf{Prefix-credit comparison.} Using the same OPTS trees and snapshots, we reconstruct max- and mean-backup return advantages with $V=0$ and compare their difference at shared prefixes. Panel~(b) reports token-weighted mean extra credit, grouped by the distance in tokens to the first downstream branch, for $s\in\{1,3,7\}$. Solid curves use $\lambda=1$ and dashed curves use $\lambda=0.999$. Thus the panel measures the credit introduced by max backup on the searched trees; it does not estimate a separate critic effect.

\subsection{Cross-Domain Policy Learning}
\label{app:setup-rq3}

\noindent\textbf{MuJoCo.} The suite contains \texttt{Hopper-v4}, \texttt{Walker2d-v4}, \texttt{HalfCheetah-v4}, \texttt{Ant-v4}, and \texttt{Humanoid-v4}. PPO and OPTS-TTPO use one million environment steps and ten random seeds per task. OPTS-TTPO uses max-backup TreeGAE and action-level performance-difference selection with $(\xi,S_{\max})=(0.6,1)$.

\noindent\textbf{Atari-57.} We use \texttt{NoFrameskip} variants with sticky actions ($p_{\text{repeat}}=0.25$ per emulated frame), agent frame skip 4, frame max-pooling, $84\times84$ grayscale observations, four-frame stacking, reward clipping, and up to 30 no-ops on reset. CleanRL-style PPO uses a CNN actor--critic. Each method is trained for 10M environment steps with eight parallel environments, 128-step rollouts, and three random seeds per game. OPTS-TTPO reuses $(\xi,S_{\max})=(0.6,1)$, uses mean-backup TreeGAE, and selects states with the action-level performance-difference estimate. We summarize each run by the mean return over all logged points and over the final 100 logged points; IQM is computed over the 57 games and three seeds after human normalization, and task wins compare the three-seed mean within each game.

\noindent\textbf{LLM RLVR.} We train four Qwen3 models with VeRL for 400 steps on \texttt{math12k} and the competition subset of \texttt{NuminaMath-1.5-RL-Verifiable} (16K prompts), comparing OPTS-TTPO with PPO, DAPO, and REINFORCE++ at 4,096 completed rollouts per update. OPTS-TTPO uses eight rounds of 512 trajectories. At each selection step, $p=0.3$ caps the fraction of established trees that may enter search. All four models use the same maximum search count $S_{\max}=3$ for OPTS-TTPO training. Training uses actor/critic learning rates $10^{-6}/10^{-5}$, max-backup TreeGAE with $\lambda=0.999$, and the zero baseline. The answer verifier assigns reward 1 to a verified correct final answer and 0 otherwise; intermediate response-token rewards are zero, and the terminal response token receives this verifier reward, which TreeGAE propagates to preceding tokens. Evaluation uses \texttt{MATH500} (the \texttt{math12k} test split), \texttt{MinervaMath}, \texttt{AMC23}, \texttt{AIME24}, \texttt{AIME25}, and \texttt{AIME26}, containing 500, 272, 40, 30, 30, and 30 problems. The main table reports both equal-weight Macro Average and question-pooled Micro Average; final comparisons use step 400 and 32 independent responses per prompt, sampled at temperature 1.0 and top-$p=0.95$, with top-$k=-1$ (no truncation) and a maximum response length of 2,048 tokens.

\subsection{Optimization Hyperparameters}
\label{app:optimization-hyperparameters}

Within each control domain, PPO and OPTS-TTPO use the same optimizer and update settings; they differ only in tree construction, advantage backup, and branch weighting. Table~\ref{tab:control-optimization-hyperparameters} lists the complete shared settings.

\begin{table}[!htbp]
\centering
\caption{Optimization hyperparameters shared by PPO and OPTS-TTPO in the control experiments.}
\label{tab:control-optimization-hyperparameters}
\small
\renewcommand{\arraystretch}{1.08}
\begin{tabular}{@{}>{\raggedright\arraybackslash}p{0.28\textwidth}>{\raggedright\arraybackslash}p{0.32\textwidth}>{\raggedright\arraybackslash}p{0.32\textwidth}@{}}
\toprule
Setting & MuJoCo & Atari-57 \\
\midrule
Optimizer & Adam, $\epsilon=10^{-5}$ & Adam, $\epsilon=10^{-5}$ \\
Learning rate & $3\times10^{-4}$, linearly annealed to 0 & $2.5\times10^{-4}$, linearly annealed to 0 \\
Rollout batch & $1\times2048=2048$ transitions & $8\times128=1024$ transitions \\
Minibatches / update epochs & 32 (64 transitions each) / 10 & 4 (256 transitions each) / 4 \\
Discount / GAE parameter & $\gamma=0.99$, $\lambda=0.95$ & $\gamma=0.99$, $\lambda=0.95$ \\
Policy clip / clipped value loss & 0.2 / yes & 0.1 / yes \\
Entropy / value-loss coefficient & 0 / 0.5 & 0.01 / 0.5 \\
Advantage normalization / gradient norm & yes / 0.5 & yes / 0.5 \\
Independent training seeds & 10 per task and method & 3 per game and method \\
\bottomrule
\end{tabular}
\end{table}

All LLM methods use AdamW with $\beta=(0.9,0.999)$, $\epsilon=10^{-8}$, weight decay 0.1, actor learning rate $10^{-6}$, ten warmup steps followed by a constant learning rate, gradient clipping at 1.0, one policy-optimization epoch per update, zero entropy coefficient, and no KL penalty in either the reward or loss. Training rollouts use temperature 1.0, top-$p=1.0$, no top-$k$ truncation, and a maximum response length of 2,048 tokens. Table~\ref{tab:llm-optimization-hyperparameters} gives the method-specific settings.

\begin{table}[!htbp]
\centering
\caption{Method-specific hyperparameters for LLM RLVR.}
\label{tab:llm-optimization-hyperparameters}
\small
\renewcommand{\arraystretch}{1.08}
\begin{tabular}{@{}>{\raggedright\arraybackslash}p{0.16\textwidth}>{\raggedright\arraybackslash}p{0.76\textwidth}@{}}
\toprule
Method & Method-specific settings \\
\midrule
PPO & 4,096 prompts and rollouts per update; GAE with $\gamma=1$ and $\lambda=0.999$; learned critic with learning rate $10^{-5}$ and value clip 0.5; symmetric policy clip 0.2; token-level loss aggregation. \\
DAPO & 512 prompts with 8 responses each; group-relative advantages and no critic; asymmetric policy clip $(0.2,0.28)$ with dual-clip constant 10; token-level loss aggregation; overlong-response buffer of 1,024 tokens with penalty factor 1.0. \\
REINFORCE++ & 512 prompts with 8 responses each; per-prompt mean reward baseline followed by global advantage whitening, with no critic; symmetric policy clip 0.2; token-level loss aggregation. \\
OPTS-TTPO & 4,096 rollouts collected over the search rounds specified in Appendix~\ref{app:experiment-setup}; TreeGAE with a learned critic at learning rate $10^{-5}$; symmetric policy clip 0.2; branch-weighted token-level loss aggregation. \\
\bottomrule
\end{tabular}
\end{table}

\subsection{Simulator State Restoration}
\label{app:simulator-restoration}

OPTS-TTPO uses simulator snapshot-and-restore access to sample new suffixes from visited states, while PPO collects ordinary forward rollouts. Interaction budgets count newly generated transitions, including branch transitions; restoring a snapshot does not itself advance the environment or count as an interaction.

\noindent\textbf{MuJoCo.} We snapshot joint positions, velocities, simulation time, auxiliary simulator fields, and derived quantities used by observations. Snapshots also preserve the environment random-number-generator state, elapsed episode steps, episode statistics, observation/reward normalization statistics, and the discounted-return accumulator. Restoration reinstates the physical state, calls \texttt{mj\_forward}, and restores the saved derived quantities and wrapper state. Each environment is restored independently to the selected state and its cached observation before new actions are sampled from the current policy.

\noindent\textbf{Atari.} We use ALE's \texttt{cloneSystemState} and \texttt{restoreSystemState} to save and restore emulator state, including the sticky-action random-number-generator state. Snapshots also preserve frame-stack and max-pooling buffers, life-loss bookkeeping, elapsed-step and episode-statistics counters, and the random state used for reset no-ops. Thus a restored branch resumes the saved environment random stream, while actions are newly sampled from the current policy. Environments maintain separate snapshots, with batched policy inference. After a terminal or truncated trajectory, a selected branch restores its snapshot and observation; otherwise, the environment follows the standard wrapped reset, including episodic-life handling. Each update collects 128 new transitions per environment, bootstrapping nonterminal rollout boundaries with the critic. Tree records are rebuilt after each policy update; unfinished environment trajectories continue, but their carried-over initial segments are excluded from branching.

\subsection{Training--Evaluation Overlap Audit}
\label{app:data-overlap-audit}

Before constructing the 16,384-prompt LLM training split, we audit the union of the \texttt{math12k} and \texttt{NuminaMath-1.5-RL-Verifiable} training data against all 902 problems in \texttt{MATH500}, \texttt{MinervaMath}, \texttt{AMC23}, and \texttt{AIME24--26}. For each example, we extract the user problem, remove all whitespace, and convert it to lowercase. We remove every training example whose normalized problem exactly matches any normalized evaluation problem, as well as duplicate normalized problems within the merged training pool. We then apply the 1,024-token prompt-length filter, shuffle the remaining pool, and retain 16,384 prompts. The resulting training split has zero exact normalized matches with the six evaluation sets. This audit detects exact problem reuse up to whitespace and letter case; it does not treat paraphrases as matches.

\subsection{LLM Prompt and Chat Template}
\label{app:llm-prompt-template}

All LLM training and evaluation examples use the same system prompt:

\begin{center}
\begin{tcolorbox}[width=0.95\textwidth,title=Instruction Prompt for Mathematical Reasoning]
You are a math problem solver. For each problem, think through it step by step within \texttt{\textless think\textgreater{} \textless/think\textgreater{}} tags, then provide your final answer using \texttt{\textbackslash boxed\{\}}.\\

Requirements:\\
- Show your complete reasoning process inside \texttt{\textless think\textgreater{}} tags.\\
- Provide your final answer inside \texttt{\textbackslash boxed\{\}}.\\

Example:\\
User: If 3x + 7 = 22, what is x?\\
Assistant: \texttt{\textless think\textgreater{}}\\
3x + 7 = 22\\
3x = 22 - 7 = 15\\
x = 15 / 3 = 5\\
\texttt{\textless/think\textgreater{}}\\
The answer is \texttt{\textbackslash boxed\{5\}}.
\end{tcolorbox}
\end{center}

For every dataset, the original problem text is inserted without an additional task-specific instruction using the following message structure:

\begin{verbatim}
[
  {"role": "system", "content": SYSTEM_PROMPT},
  {"role": "user",   "content": PROBLEM}
]
\end{verbatim}

We do not supply a custom Jinja template. VeRL applies the chat template distributed with the corresponding Qwen3 checkpoint with \texttt{add\_generation\_prompt=True}; thus the two messages are serialized with the checkpoint's system, user, and assistant role delimiters, followed by the assistant generation prefix. Qwen3 thinking mode is retained: evaluation sets \texttt{enable\_thinking=True} explicitly, while training uses the checkpoint template's default thinking behavior. The same message construction and tokenizer template are used for all four training methods and all six evaluation sets.

\subsection{LLM Reasoning as a Markov Decision Process}
\label{app:llm-mdp}

The initial state $s_o\sim\rho_0$ is a problem prompt. At generation step $t$, the state contains the prompt and the generated response prefix, and the action is the next response token:
\[
s_t=(s_o,a_0,\ldots,a_{t-1}),
\qquad a_t\sim\pi_\theta(\cdot\mid s_t).
\]
The transition appends $a_t$ to the prefix deterministically. An episode ends at an end-of-sequence token or the response-length limit. Intermediate rewards are zero; the final response token receives the answer-verification score. The critic estimates the remaining return from each prefix, and TreeGAE propagates the terminal reward through the sampled suffixes. Its regression target is $\widehat R_x=\widehat A_x+V_{\bar\phi}(s_x)$, as in Section~\ref{sec:ttpo-objective}.

Rebranching at $s_t$ retains the prompt and response prefix and samples a new suffix from $\pi_\theta$. Thus each transition occurrence corresponds to a response-token occurrence, and multiple suffixes share their sampled prefix. OPTS selects rebranching positions using the backed-up scores; TTPO aggregates the resulting token contributions using the branch weights $W(x)$.

\section{Rollout Scaling Across Individual Benchmarks}
\label{app:rq2-compute-scaling-by-dataset}

\begin{figure*}[!htbp]
\centering
\includegraphics[width=\linewidth]{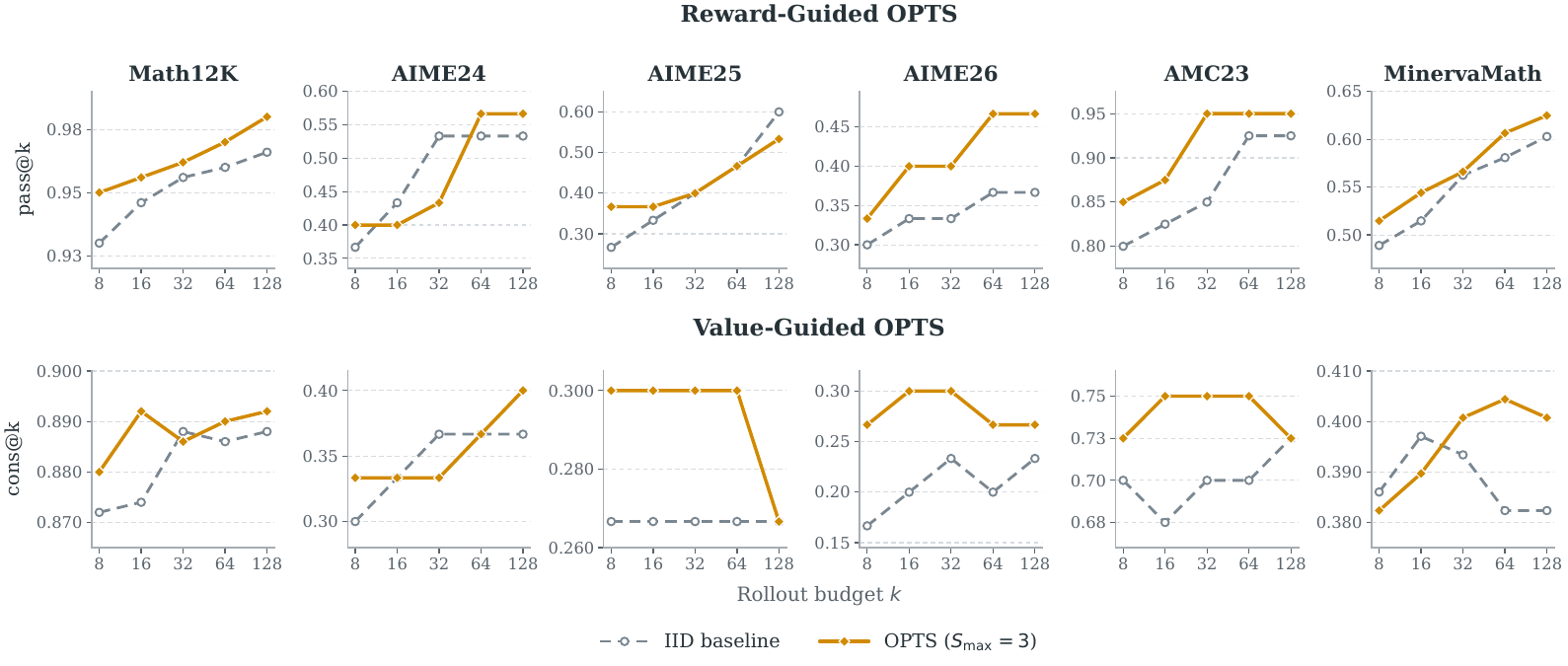}
\caption{\textbf{Rollout scaling of OPTS across individual reasoning benchmarks.} Top: reward-guided OPTS versus i.i.d. \texttt{pass@k}. Bottom: value-guided OPTS versus i.i.d. self-consistency. Each column reports one benchmark under matched rollout budgets $k\in\{8,16,32,64,128\}$ at fixed $S_{\max}=3$.}
\label{fig:rq2-compute-scaling-by-dataset}
\end{figure*}
\FloatBarrier

\section{Search-Budget Improvement of OPTS}
\label{app:search-budget-improvement}

\begin{figure*}[!htbp]
\centering
\includegraphics[width=\linewidth]{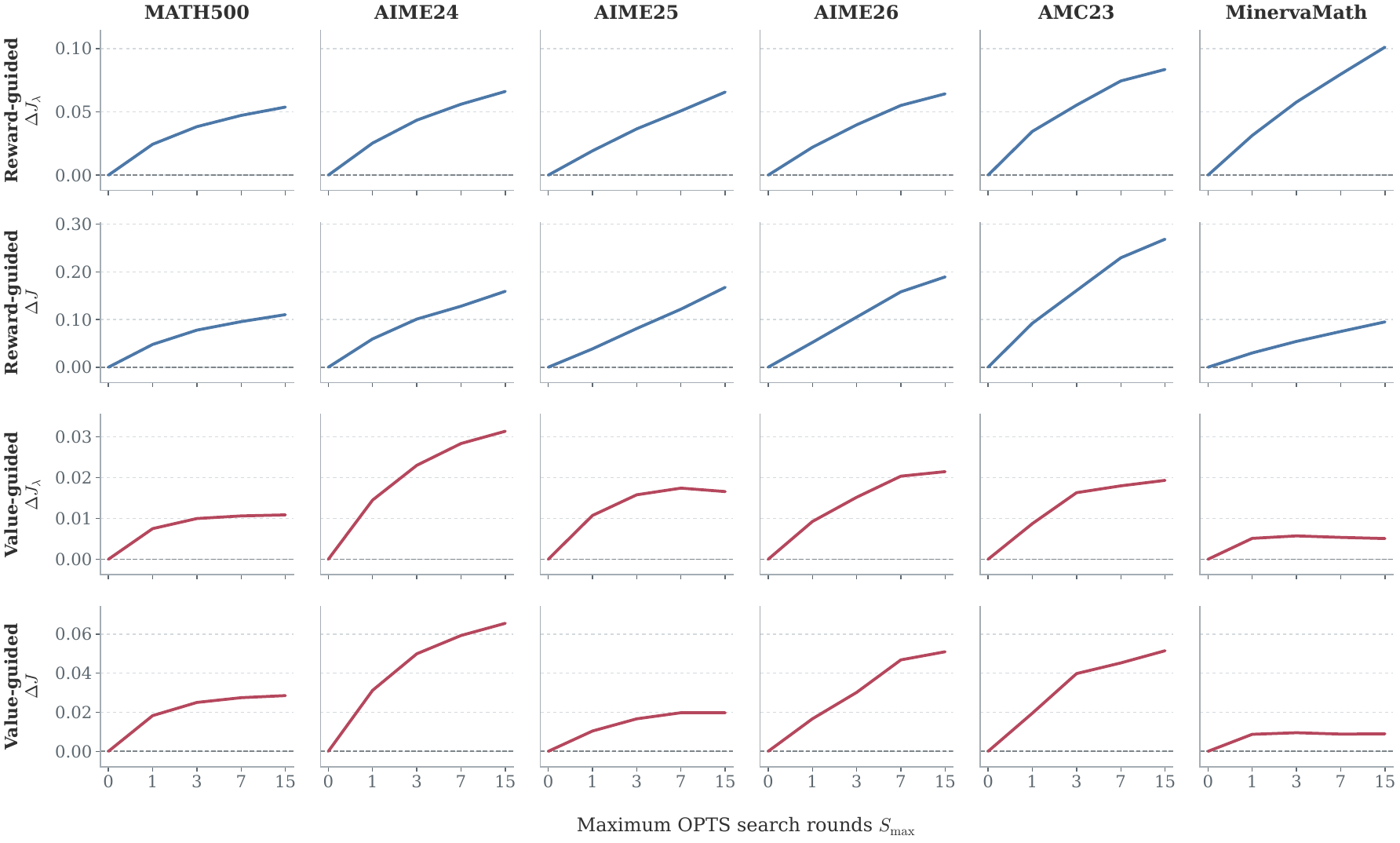}
\caption{\textbf{Learned-critic search-budget results by benchmark.} Rows report reward-guided $\Delta J_\lambda$, reward-guided $\Delta J$, value-guided $\Delta J_\lambda$, and value-guided $\Delta J$, where each difference is relative to $S_{\max}=0$. Columns show the six reasoning benchmarks.}
\label{fig:rq2-search-budget-scaling}
\end{figure*}

Theorem~2 assumes a fixed exact value function; this experiment tests the trend with a learned critic. We fix the step-400 \texttt{Qwen3-1.7B} OPTS-TTPO policy and critic, set $\lambda=0.999$, and vary $S_{\max}\in\{0,1,3,7,15\}$ with 32 trees per prompt on 902 held-out prompts. Reward- and value-guided search reuse the same root-response texts. We report the greedy-path verifier $\lambda$-return $J_\lambda$ and selected-terminal return $J$, each relative to $S_{\max}=0$. Figure~\ref{fig:rq2-search-budget-scaling} gives all four quantities by benchmark. Reward-guided curves increase throughout; value-guided $J_\lambda$ peaks before $S_{\max}=15$ on \texttt{AIME25} and \texttt{MinervaMath}, so the exact-value guarantee need not hold with an approximate critic. Micro averages nevertheless increase over all five budgets, by $(0.0706,0.1190)$ for reward guidance and $(0.0107,0.0254)$ for value guidance in $(J_\lambda,J)$. The reward-guided gain also depends on rebranching position (Appendix~\ref{app:rebranch-position-ablation}).

\FloatBarrier

\section{Rebranching Position: Performance-Difference versus Random and Midpoint Selection}
\label{app:rebranch-position-ablation}

\begin{figure*}[!htbp]
\centering
\includegraphics[width=\linewidth]{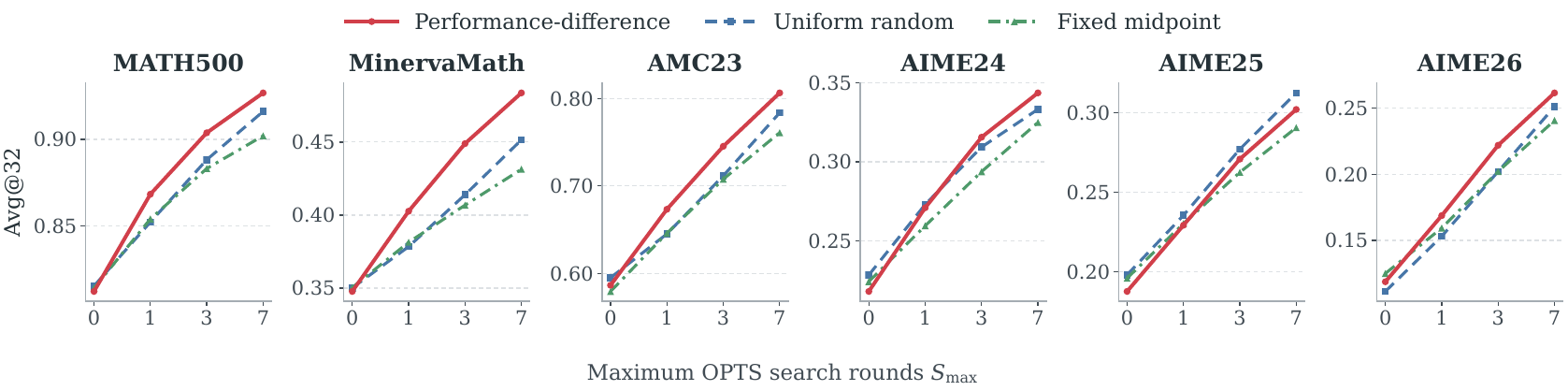}
\caption{\textbf{Rebranching position matters.} \texttt{Avg@32} versus the maximum number of OPTS search rounds for performance-difference, uniformly random, and fixed-midpoint rebranching on the same trees and budgets.}
\label{fig:rebranch-position-ablation}
\end{figure*}

OPTS always rebranches by the performance-difference estimate, so we test whether the position it picks matters. Using the step-400 \texttt{Qwen3-1.7B} OPTS-TTPO policy and critic, 902 held-out prompts, and 32 trees per prompt, we run OPTS under three rebranching rules: performance-difference selection, a uniformly random position on the same greedy path, and the fixed midpoint of that path. The two control rules keep the per-round tree selection of OPTS, so the number of rebranchings per round is matched; only the node position changes. Figure~\ref{fig:rebranch-position-ablation} reports the greedy-terminal \texttt{Avg@32} as the search budget grows. Performance-difference selection is best on five of the six benchmarks at every budget above zero, and its margin widens with $S_{\max}$; on AIME25 random selection is marginally higher.

\FloatBarrier

\section{MuJoCo Hyper-Parameter Grid: Length-Penalty Exponent and Search Count}
\label{app:mujoco-xi-s-grid}

\begin{figure*}[!htbp]
\centering
\includegraphics[width=\linewidth]{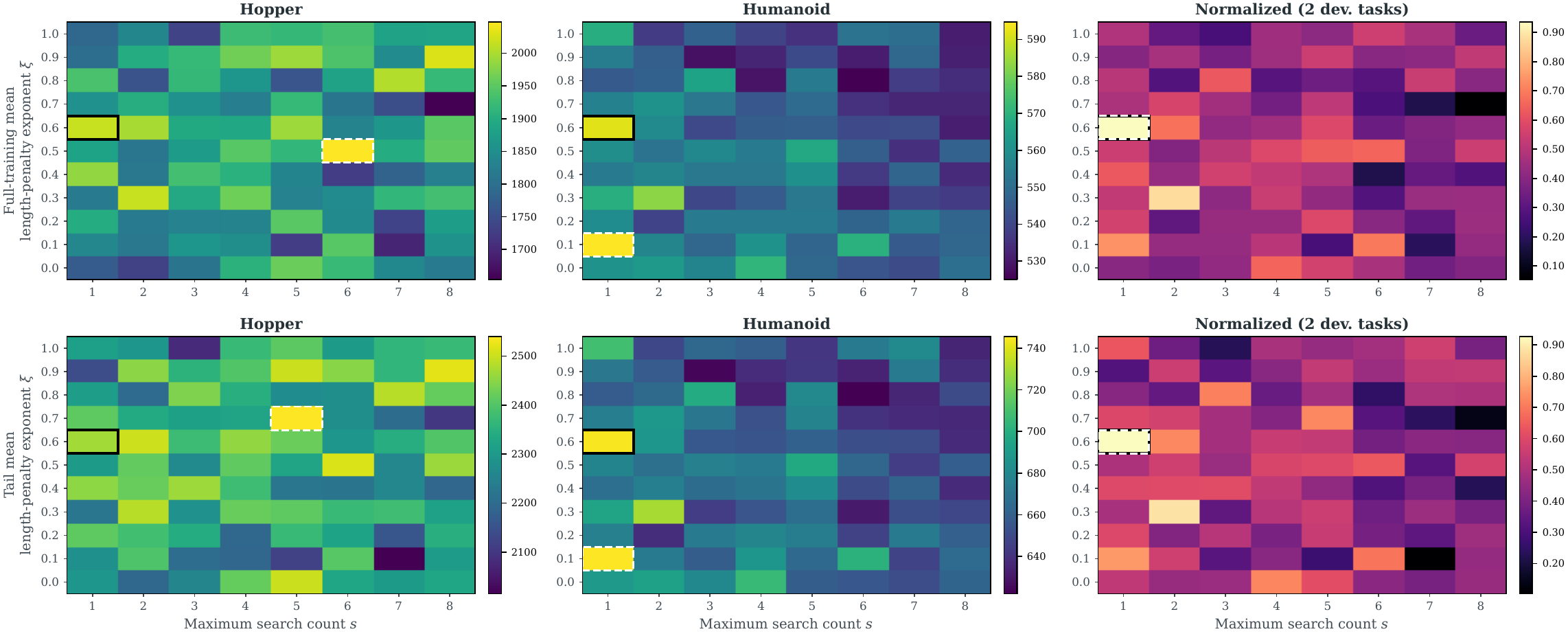}
\caption{\textbf{$\xi\times s$ grid on the MuJoCo development tasks.} Top: full-training mean return; bottom: tail return. Black outline: the selected configuration $(0.6,1)$; white dashed outline: the best cell of each panel. The right-most column averages the min--max-normalized grids of Hopper-v4 and Humanoid-v4.}
\label{fig:mujoco-xi-s-grid}
\end{figure*}

We sweep $\xi\in\{0,0.1,\ldots,1.0\}$ and $s\in\{1,\ldots,8\}$ on Hopper-v4 and Humanoid-v4, the smallest- and largest-action-space tasks in the five-task suite. The remaining OPTS-TTPO settings are unchanged: 1M steps, 10 seeds per cell, 88 cells per task, and max-backup TreeGAE. Figure~\ref{fig:mujoco-xi-s-grid} reports full-training and tail returns plus a cross-task summary that min--max normalizes each grid before averaging. The main-text setting $(\xi,s)=(0.6,1)$ (black outline) is best under both metrics and is fixed for the remaining tasks; the heatmaps show the complete grid.

\FloatBarrier

\section{Full Atari-57 Learning Curves}
\label{app:atari-curves}

Each panel corresponds to one Atari game, the horizontal axis is the number of environment interaction steps, and the vertical axis is the raw mean return for that game. Both methods are trained under the sticky-action protocol, and OPTS-TTPO uses mean-backup TreeGAE, the stochastic-environment setting, with $\xi=0.6$ and $S_{\max}=1$ as in the main text. Different random seeds of the same algorithm are shown as thin curves with the same color, with smoothing used only for visualization.

\begin{figure*}[p]
\centering
\includegraphics[width=\linewidth,height=0.92\textheight]{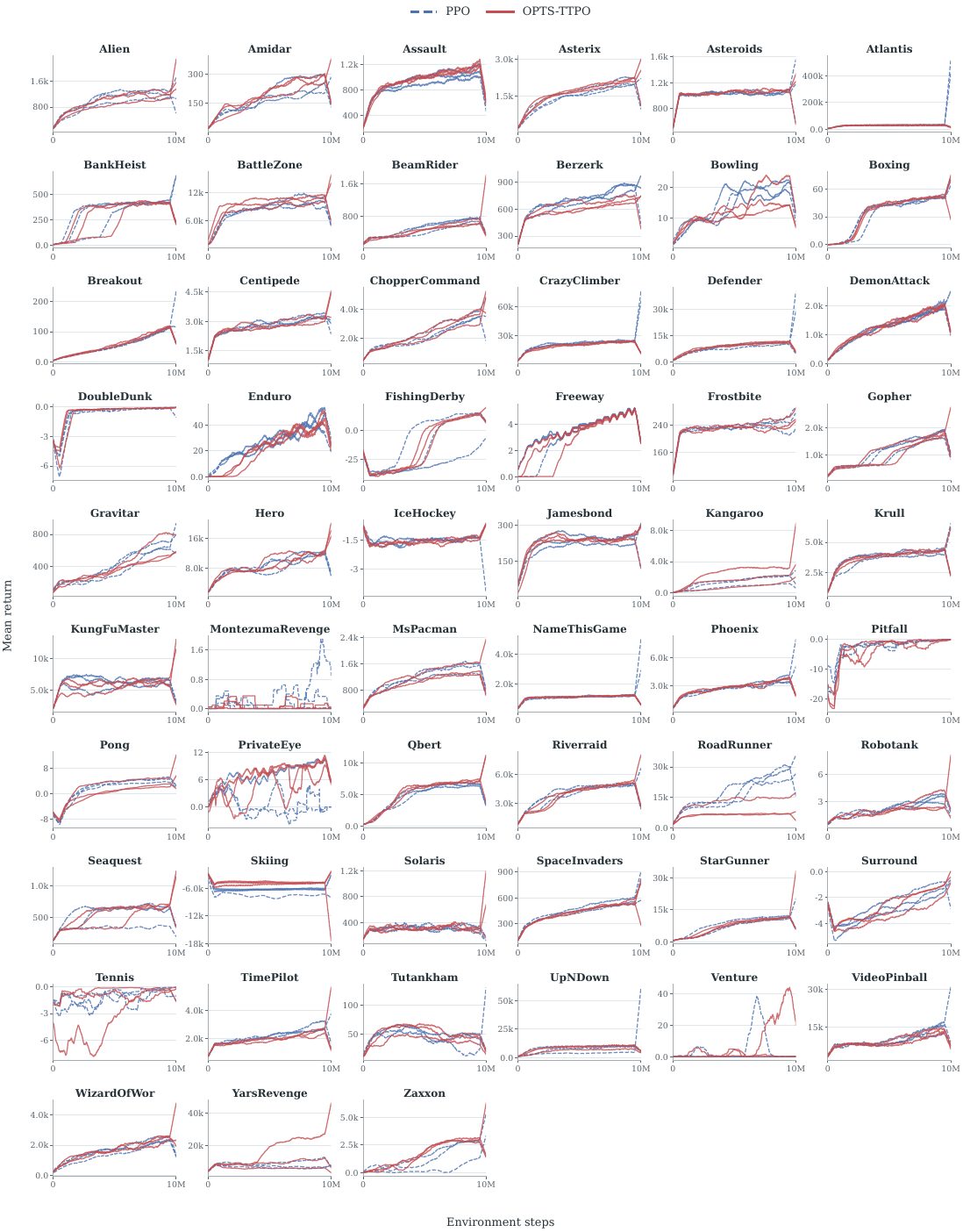}
\caption{Full Atari-57 learning curves for PPO and OPTS-TTPO. Each panel shows one game; curves are smoothed raw mean returns for different random seeds.}
\label{fig:atari-all-curves}
\end{figure*}

Because reward scales differ substantially across Atari games, Figure~\ref{fig:atari-all-curves} is intended for within-task comparison and for inspecting where gains or regressions occur. The aggregate cross-task conclusion is therefore based on the human-normalized IQM and task-level win counts reported in the main text rather than on visually aggregating these raw-return curves.

\clearpage

\section{Backup-Rule Comparison on Sticky-Action Atari}
\label{app:atari-backup-ablation}
We compare mean- and max-backup OPTS-TTPO at the same $(\xi,S_{\max})=(0.6,1)$ under the main-text sticky-action protocol: 57 games, three seeds, and ten million interaction steps per run. For each algorithm, we first average the three seed curves pointwise, then average all logged points or the final 100 points. Win/loss/tie counts compare these game-level means using an absolute tie tolerance of $10^{-12}$; the per-game tables show the same values rounded to two decimals.

\begin{table}[!htbp]
\centering
\caption{Backup-rule comparisons at fixed hyperparameters.}
\small
\begin{tabular}{@{}llrrr@{}}\toprule Metric & Comparison & Wins & Losses & Ties \\ \midrule
Full training & Max vs. Mean & 19 & 22 & 16 \\
Full training & Max vs. PPO & 27 & 30 & 0 \\
Full training & Mean vs. PPO & \textbf{31} & 26 & 0 \\
Tail & Max vs. Mean & 18 & 22 & 17 \\
Tail & Max vs. PPO & \textbf{34} & 22 & 1 \\
Tail & Mean vs. PPO & \textbf{34} & 22 & 1 \\
\bottomrule\end{tabular}
\end{table}

Mean backup has a slight edge in game-level wins against max backup, while both beat PPO on 34 games under the tail metric. These results do not establish universal superiority of either backup rule or isolate environmental stochasticity as a causal factor.

\FloatBarrier
\begin{table}[!htbp]
\centering
\caption{Per-game seed-mean returns (1), shown to two decimals. Full: all logged points; Tail: final 100 points. Win counts use unrounded values.}
\scriptsize
\begin{tabular}{@{}lrrrrrr@{}}\toprule
& \multicolumn{3}{c}{Full} & \multicolumn{3}{c}{Tail}\\
Game & PPO & Mean & Max & PPO & Mean & Max \\ \midrule
ALE\_Surround-v5 & -2.83 & -2.94 & -2.94 & -1.14 & -0.91 & -0.91 \\
AlienNoFrameskip-v4 & 989.13 & 947.58 & 945.15 & 1186.07 & 1236.76 & 1372.65 \\
AmidarNoFrameskip-v4 & 173.34 & 192.09 & 191.86 & 246.73 & 301.20 & 307.77 \\
AssaultNoFrameskip-v4 & 922.34 & 967.38 & 968.90 & 1073.07 & 1278.37 & 1208.08 \\
AsterixNoFrameskip-v4 & 1553.26 & 1671.41 & 1690.02 & 2051.14 & 2223.03 & 2194.67 \\
AsteroidsNoFrameskip-v4 & 1031.89 & 1045.94 & 1042.91 & 1077.44 & 1081.97 & 1105.55 \\
AtlantisNoFrameskip-v4 & 30513.96 & 30637.19 & 28223.74 & 40791.67 & 32212.33 & 27254.17 \\
BankHeistNoFrameskip-v4 & 310.09 & 288.34 & 221.75 & 430.11 & 437.08 & 406.26 \\
BattleZoneNoFrameskip-v4 & 8426.72 & 9203.13 & 9694.18 & 10452.50 & 10219.44 & 11091.39 \\
BeamRiderNoFrameskip-v4 & 487.94 & 451.52 & 451.52 & 704.01 & 663.40 & 663.40 \\
BerzerkNoFrameskip-v4 & 688.99 & 612.03 & 618.03 & 836.57 & 714.46 & 735.34 \\
BowlingNoFrameskip-v4 & 14.40 & 12.27 & 12.39 & 18.25 & 16.85 & 17.46 \\
BoxingNoFrameskip-v4 & 33.90 & 35.09 & 35.09 & 51.22 & 54.24 & 54.24 \\
BreakoutNoFrameskip-v4 & 55.06 & 58.26 & 54.77 & 120.19 & 125.61 & 115.80 \\
CentipedeNoFrameskip-v4 & 2818.38 & 2761.23 & 2806.74 & 3237.38 & 3071.08 & 3212.12 \\
ChopperCommandNoFrameskip-v4 & 2229.34 & 2249.77 & 2135.29 & 3819.40 & 3684.47 & 3281.21 \\
CrazyClimberNoFrameskip-v4 & 21412.02 & 20278.52 & 20278.52 & 23151.00 & 23777.56 & 23777.56 \\
DefenderNoFrameskip-v4 & 8376.84 & 8961.33 & 8783.60 & 10542.53 & 12488.08 & 12195.61 \\
DemonAttackNoFrameskip-v4 & 1276.39 & 1302.78 & 1314.93 & 2044.33 & 1925.20 & 2188.04 \\
DoubleDunkNoFrameskip-v4 & -0.81 & -0.70 & -0.70 & -0.05 & -0.06 & -0.06 \\
EnduroNoFrameskip-v4 & 26.72 & 22.69 & 22.69 & 40.72 & 33.14 & 33.14 \\
FishingDerbyNoFrameskip-v4 & -16.08 & -12.51 & -12.51 & 5.97 & 14.45 & 14.45 \\
FreewayNoFrameskip-v4 & 3.52 & 3.24 & 3.24 & 5.96 & 5.88 & 5.88 \\
FrostbiteNoFrameskip-v4 & 231.70 & 231.77 & 237.73 & 250.58 & 249.40 & 255.50 \\
GopherNoFrameskip-v4 & 1122.37 & 1075.35 & 904.68 & 1912.06 & 1894.34 & 1332.87 \\
GravitarNoFrameskip-v4 & 397.21 & 376.87 & 336.78 & 642.40 & 617.09 & 518.12 \\
HeroNoFrameskip-v4 & 8702.10 & 8840.72 & 9149.41 & 11663.23 & 12233.24 & 12438.86 \\
IceHockeyNoFrameskip-v4 & -1.54 & -1.60 & -1.58 & -1.38 & -1.28 & -1.19 \\
JamesbondNoFrameskip-v4 & 223.91 & 229.30 & 234.10 & 248.14 & 267.07 & 263.86 \\
\bottomrule\end{tabular}
\end{table}
\begin{table}[!htbp]
\centering
\caption{Per-game seed-mean returns (2), shown to two decimals. Full: all logged points; Tail: final 100 points. Win counts use unrounded values.}
\scriptsize
\begin{tabular}{@{}lrrrrrr@{}}\toprule
& \multicolumn{3}{c}{Full} & \multicolumn{3}{c}{Tail}\\
Game & PPO & Mean & Max & PPO & Mean & Max \\ \midrule
KangarooNoFrameskip-v4 & 1229.65 & 1612.89 & 1587.39 & 1802.94 & 2458.78 & 2444.39 \\
KrullNoFrameskip-v4 & 3762.75 & 3802.40 & 3776.63 & 4173.87 & 4482.65 & 4402.41 \\
KungFuMasterNoFrameskip-v4 & 6102.91 & 5726.26 & 5726.26 & 5609.22 & 6604.17 & 6604.17 \\
MontezumaRevengeNoFrameskip-v4 & 0.21 & 0.06 & 0.48 & 0.00 & 0.17 & 0.00 \\
MsPacmanNoFrameskip-v4 & 1185.86 & 1131.68 & 1184.35 & 1471.88 & 1464.83 & 1551.24 \\
NameThisGameNoFrameskip-v4 & 1160.37 & 1182.86 & 1182.86 & 1337.90 & 1352.00 & 1352.00 \\
PhoenixNoFrameskip-v4 & 2744.98 & 2826.81 & 2768.06 & 3813.27 & 3939.31 & 3836.27 \\
PitfallNoFrameskip-v4 & -3.08 & -3.89 & -4.91 & 0.00 & 0.00 & -0.48 \\
PongNoFrameskip-v4 & 1.54 & 0.58 & 0.58 & 4.61 & 3.50 & 3.50 \\
PrivateEyeNoFrameskip-v4 & 2.80 & 6.12 & 6.11 & 4.67 & 12.20 & 12.20 \\
QbertNoFrameskip-v4 & 4894.80 & 5279.52 & 4903.30 & 6714.89 & 7282.50 & 7130.39 \\
RiverraidNoFrameskip-v4 & 4081.11 & 3991.97 & 4060.80 & 4840.25 & 5520.78 & 4899.51 \\
RoadRunnerNoFrameskip-v4 & 17287.81 & 8471.32 & 20939.21 & 28086.25 & 9731.23 & 33341.33 \\
RobotankNoFrameskip-v4 & 2.23 & 2.02 & 2.02 & 3.56 & 2.88 & 2.88 \\
SeaquestNoFrameskip-v4 & 498.59 & 526.78 & 364.72 & 582.07 & 709.10 & 399.03 \\
SkiingNoFrameskip-v4 & -6673.30 & -4877.35 & -4877.35 & -6523.64 & -4600.00 & -4600.00 \\
SolarisNoFrameskip-v4 & 304.76 & 308.64 & 308.64 & 293.27 & 295.07 & 295.07 \\
SpaceInvadersNoFrameskip-v4 & 442.97 & 430.26 & 424.46 & 606.18 & 568.85 & 496.05 \\
StarGunnerNoFrameskip-v4 & 7459.37 & 7351.37 & 7403.09 & 11215.28 & 11361.83 & 11826.67 \\
TennisNoFrameskip-v4 & -1.17 & -1.73 & -1.73 & -0.14 & -0.15 & -0.15 \\
TimePilotNoFrameskip-v4 & 2206.19 & 1999.81 & 2059.30 & 2904.39 & 2557.28 & 2662.74 \\
TutankhamNoFrameskip-v4 & 46.56 & 48.76 & 44.14 & 37.73 & 40.55 & 48.30 \\
UpNDownNoFrameskip-v4 & 6777.72 & 7977.87 & 6655.13 & 9677.93 & 11131.87 & 8754.43 \\
VentureNoFrameskip-v4 & 1.71 & 3.62 & 3.62 & 0.00 & 16.50 & 16.50 \\
VideoPinballNoFrameskip-v4 & 9998.85 & 9542.93 & 9625.38 & 17789.33 & 15498.07 & 15813.72 \\
WizardOfWorNoFrameskip-v4 & 1602.55 & 1657.13 & 1644.41 & 2541.53 & 2613.94 & 2636.61 \\
YarsRevengeNoFrameskip-v4 & 7644.99 & 10684.72 & 8326.65 & 8370.44 & 15992.55 & 11214.26 \\
ZaxxonNoFrameskip-v4 & 1087.67 & 1727.66 & 1744.45 & 2367.19 & 3139.61 & 3122.94 \\
\bottomrule\end{tabular}
\end{table}
\FloatBarrier

\section{Per-Checkpoint LLM Evaluation Curves}
\label{app:llm-per-step}

The four \texttt{Qwen3-1.7B} methods are evaluated every 20 steps from 20 to 400 using the final-checkpoint protocol: 32 responses per prompt on the 902-prompt set, scored by the same verifier. OPTS-TTPO uses the main-text configuration with max-backup TreeGAE and $S_{\max}=3$. It matches critic-free baselines on \texttt{avg@32} for most of training and pulls ahead in the second half, especially on AIME24--26; its \texttt{pass@32} advantage appears early and persists except on AIME24.

\begin{figure}[!htbp]
\centering
\includegraphics[width=\textwidth]{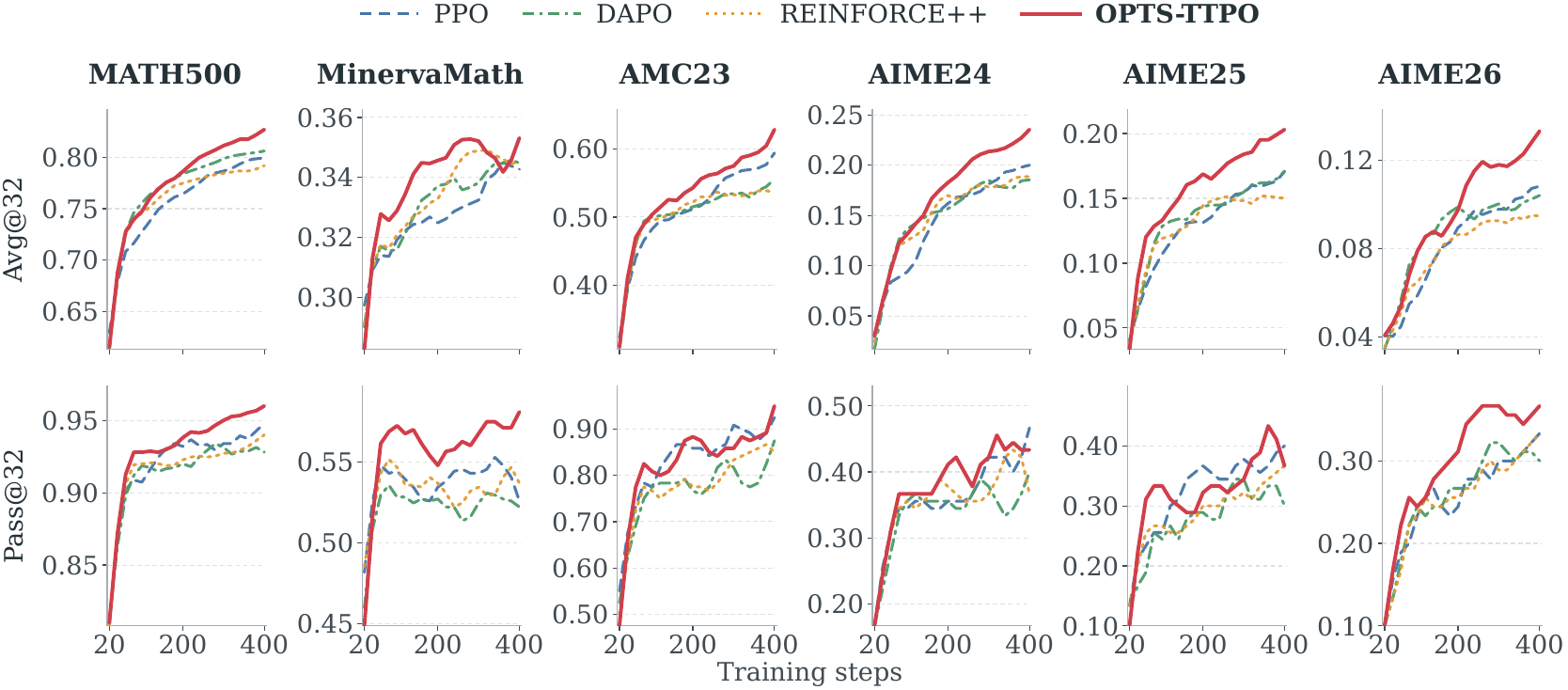}
\caption{\textbf{Qwen3-1.7B per-checkpoint evaluation across six reasoning benchmarks.} Each column shows one benchmark, with \texttt{avg@32} above \texttt{pass@32}; all 20 checkpoints of every method are scored with the same verifier. Curves show a three-point moving average that retains the step-20 and step-400 endpoints.}
\label{fig:llm-per-step-evaluation}
\end{figure}
\clearpage

\section{Wall-Clock Composition of OPTS-TTPO}
\label{app:search-time-overhead}

Table~\ref{tab:search-time-overhead} reports the internal runtime composition recorded in the \texttt{Qwen3-1.7B} OPTS-TTPO training logs for Section~\ref{sec:llm-train-search}. Rollout generation accounts for $43.7\%$ of each step, critic forwards and updates for $34.0\%$, and actor updates for $16.5\%$. The explicitly timed path refresh, branch selection, and TreeGAE backup account for $0.7\%$; beyond these operations, OPTS also makes the rollout rounds serial and performs critic forwards after each round. All timings are per training step, summing repeated operations over all rollout rounds.

\begin{table}[!htbp]
\centering
\caption{\textbf{Wall-clock composition of an OPTS-TTPO step} (\texttt{Qwen3-1.7B}, $S_{\max}=3$, eight H200 GPUs).}
\label{tab:search-time-overhead}
\small
\begin{tabular}{@{}lrr@{}}
\toprule
Stage of one training step & Seconds & Share \\
\midrule
Rollout generation (vLLM, all rounds) & 348.1 & 43.7\% \\
Critic update & 214.4 & 26.9\% \\
Actor update & 131.4 & 16.5\% \\
Critic value forwards (all rounds) & 56.5 & 7.1\% \\
Old log-probabilities & 39.2 & 4.9\% \\
\textbf{Search module: $\widehat\Delta$ path refresh + selection} & 3.2 & 0.4\% \\
\textbf{Search module: TreeGAE backup} & 2.7 & 0.3\% \\
Reward, final tree processing, other & 0.5 & 0.1\% \\
\midrule
Total & 796.0 & 100\% \\
\bottomrule
\end{tabular}
\end{table}
\FloatBarrier